\documentclass[sigconf]{acmart}

\usepackage[ruled,linesnumbered]{algorithm2e}
\usepackage{hyperref}
\usepackage{xspace}

\usepackage{amssymb}
\usepackage{framed}
\usepackage{xcolor}
\usepackage[most]{tcolorbox}
\usepackage{multirow}
\usepackage{colortbl}
\usepackage{makecell} 
\usepackage{booktabs} 
\usepackage{bbding}
\usepackage{amsmath}
\usepackage{amsthm}
\usepackage{pifont}
\usepackage[normalem]{ulem}

\usepackage{titlesec} 
\titleformat{\subsubsection}[block]  
  {\normalfont\large\bfseries}      
  {\thesubsubsection}               
  {1em}                             
  {}                                

\usepackage{framed}
\definecolor{shadecolor}{rgb}{0.92,0.92,0.92}

\usepackage{enumerate}

\definecolor{ForestGreen}{RGB}{34,139,34}

\setcopyright{none}
\renewcommand\footnotetextcopyrightpermission[1]{} 
\newcommand{\hi}[1]{\vspace{.25em} \noindent {\bf #1}\xspace}

\newcommand{\oursys}{\texttt{ST-Raptor}\xspace}

\newcommand{\zxh}[1]{\textcolor{blue}{#1}}

\newcommand{\revisiona}[1]{\textcolor{black}{#1}}
\newcommand{\revisionb}[1]{\textcolor{black}{#1}}
\newcommand{\revisionc}[1]{\textcolor{black}{#1}}

\newcommand{\llm}{\textsc{LLM}\xspace}
\newcommand{\llms}{\textsc{LLMs}\xspace}
\newcommand{\bfit}[1]{\textbf{\textit{#1}}}

\newcommand{\Semitables}{\emph{Semi-Structured Tables}\xspace}
\newcommand{\semitables}{\emph{semi-structured tables}\xspace}
\newcommand{\semitable}{{semi-structured table}\xspace}
\newcommand{\Semiqa}{\emph{Semi-Structured Table QA}\xspace}
\newcommand{\semiqa}{\emph{semi-structured table QA}\xspace}

\definecolor{darkblue}{rgb}{0, 0, 0.5}
\hypersetup{colorlinks=true, citecolor=darkblue, linkcolor=darkblue, urlcolor=darkblue}
\definecolor{lightgray}{rgb}{0.9, 0.9, 0.9}
\definecolor{darkgreen}{RGB}{50,100,0}
\definecolor{darkred}{RGB}{200, 0, 0}
\definecolor{lightred}{RGB}{250, 200, 200}
\definecolor{lightblue}{RGB}{210, 220, 250}
\definecolor{doderblue}{RGB}{30,144,255}
\definecolor{select}{RGB}{222, 235, 247}
\definecolor{unselect}{RGB}{247, 207, 206}

\newtheorem{definition}{Definition}

\begin{document}

\pagestyle{plain}
\pagenumbering{roman}

\twocolumn
\pagestyle{plain}
\title{ST-Raptor: LLM-Powered Semi-Structured Table\\ Question Answering}

\renewcommand\thesection{\arabic{section}}
\setcounter{section}{0}
\pagenumbering{arabic}
\setcounter{page}{1}
\setcounter{figure}{1}
\setcounter{table}{0}

\pagestyle{plain}
\pagenumbering{arabic}

\author{Zirui Tang}
\affiliation{
    Shanghai Jiao Tong University
    \country{}    
    \city{}
}
\email{tangzirui@sjtu.edu.cn}

\author{Boyu Niu}
\affiliation{
    Shanghai Jiao Tong University
    \country{}
    \city{}    
}
\email{nby2005@sjtu.edu.cn}

\author{Xuanhe Zhou}
\authornote{ Xuanhe Zhou is the corresponding author.}
\affiliation{
    Shanghai Jiao Tong University
    \country{}
    \city{}
}
\email{zhouxh@cs.sjtu.edu.cn}

\author{Boxiu Li}
\affiliation{
    Shanghai Jiao Tong University
    \country{}
    \city{}
}
\email{lbxhaixing154@sjtu.edu.cn}

\author{Wei Zhou}
\affiliation{
    Shanghai Jiao Tong University
    \country{}
    \city{}
}
\email{weizhoudb@sjtu.edu.cn}

\author{Jiannan Wang}
\affiliation{
    \institution{Simon Fraser University}
    \country{}
    \city{}
}
\email{jnwang@sfu.ca}

\author{Guoliang Li}
\affiliation{
    \institution{Tsinghua University}
    \country{}
    \city{}
}
\email{liguoliang@tsinghua.edu.cn}

\author{Xinyi Zhang}
\affiliation{
    \institution{Renmin University of China}
    \country{}
    \city{}
}
\email{xinyizhang.info@ruc.edu.cn}

\author{Fan Wu}
\affiliation{
    Shanghai Jiao Tong University
    \country{}
    \city{}
}
\email{fwu@cs.sjtu.edu.cn}

\setcounter{figure}{0}

\begin{teaserfigure}
  \centering
  \includegraphics[width=2\linewidth]{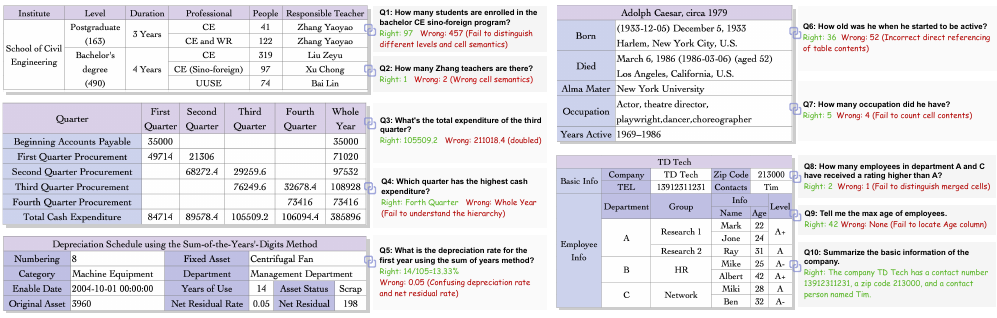}
  \caption{Example analytical questions over real-world semi-structured tables (e.g., Excel spreadsheets).}
  \label{fig:fig1}
\end{teaserfigure}

\begin{abstract}
Semi-structured tables, widely used in real-world applications (e.g., financial reports, medical records, transactional orders), often involve flexible and complex layouts (e.g., hierarchical headers and merged cells). These tables generally rely on human analysts to interpret table layouts and answer relevant natural language questions, which is costly and inefficient. To automate the procedure, existing methods face significant challenges. First, methods like NL2SQL require converting semi-structured tables into structured ones, which often causes substantial information loss. Second, methods like NL2Code and multi-modal LLM QA struggle to understand the complex layouts of semi-structured tables and cannot accurately answer corresponding questions.

To this end, we propose \oursys, a tree-based framework for semi-structured table question answering (\semiqa) using large language models. First, we introduce the Hierarchical Orthogonal Tree (HO-Tree), a structural model that captures complex \semitable layouts, along with an effective algorithm for constructing the tree by identifying headers, content values, and their implicit relationships. Second, we define a set of basic tree operations to guide \llms in executing common QA tasks. Given a user question, \oursys decomposes it into simpler sub-questions, generates corresponding tree operation pipelines, and conducts operation-table alignment for accurate pipeline execution. Third, we incorporate a two-stage verification mechanism: (1) forward validation checks the correctness of execution steps, while (2) backward validation evaluates answer reliability by reconstructing queries from predicted answers. To benchmark the performance, we present SSTQA, a dataset of 764 questions over 102 real-world semi-structured tables. Experiments show that \oursys outperforms nine baselines by up to {20\%} in answer accuracy. The code is available at \texttt{\textcolor{blue}{\url{https://github.com/weAIDB/ST-Raptor}}}.
\end{abstract}

\pagestyle{plain}
\pagenumbering{arabic}

\maketitle
\setcounter{table}{0}

\section{Introduction}
\label{sec:intro}


\Semitables are a type of data structure that represents the flexible and complex layouts commonly found in real-world data across a variety of applications, such as Word Tables for financial reports~\cite{FinanceData}, Excel spreadsheets for medical records~\cite{MedicalData}, and PDF Tables for e-commerce transaction orders~\cite{CommerceData}. They often serve as the major type in these applications, e.g., accounting for up to 80\% of patient records in Electronic Medical Record (EMR) systems~\cite{MedicalData}. 

For better understanding, in Figure~\ref{fig:fig1}, we showcase five example \semitables with corresponding user questions from diverse scenarios (e.g., human resource management, financial management, and personal information). Considering the bottom-right table (\texttt{TD-Tech}): $(i)$ the top portion covers the company’s fundamental details, while $(ii)$ the lower portion contains basic information and performance ratings of employees per department. Different shades of blue in \texttt{TD-Tech} highlight the nested levels of the table, reflecting relationships like hierarchical headers (e.g., the ``Basic Info'' header linked to lower-level headers like ``Company'' and ``TEL'') and header-to-content (e.g., both ``Department'' and ``Level'' headers own the content values of ``A''s). Even human analysts may need to carefully analyze the layout characters to fully understand such \semitables.


This layout flexibility makes \semiqa extremely challenging and distinguishes it from other common QA tasks on structured data (e.g., relational tables~\cite{StructuredData}) and unstructured data (e.g., textual documents or multimedia files \cite{barboule2025survey}). In Figure~\ref{fig:fig1}, we showcase \semitable questions that commonly require the following analysis strategies: $(i)$ Identify the headers based on the user question to locate the areas of relevant table cells. For instance, $Q8$ first identifies the ``Level'' header in the ``Employee Info'' sub-table, and then recognizes the ``A+'' cells. Meanwhile, it is essential to distinguish that the content value ``A'' under the ``Department'' header is different from ``Level A'' in the original question. $(ii)$ Leverage the identified cells and the original question to analyze the surrounding table structures for capturing nested relationships and extracting additional information. For instance, for $Q9$, the merged cell ``A+'' applies to two employees to get the right answer ``2''. $(iii)$ Explore all potentially relevant header and content cells required to form an accurate answer. For instance, for $Q10$, relevant information about the company is needed for summarization.

\begin{sloppypar}
Existing works (including powerful \llms like GPT4-o\cite{openai2024gpt4o} and DeepSeek-R1 \cite{deepseekr1}) face significant limitations in \semiqa. 
First, NL2SQL methods~\cite{OpenSearchSQL,XiYanSQL,ChenRTS,BeforeGeneration,PICARD,BorchmannQnC} generate SQL statements executed on relational tables to get the final answer. However, it requires converting semi-structured tables into fully structured formats, causing substantial information loss. Second, methods like NL2Code \cite{zan2023nl2code} generate python code to operate on pandas dataframes. However, they struggle to understand many complex semi-structured tables and conduct precise information retrieval. A more promising approach involves converting tables into images for processing with Vision Language Models (VLM)~\cite{TableLLaVA,DocOwl1.5}. But it has three main limitations: $(i)$ Table2image causes precise loss and often misleads to irrelevant table areas; $(ii)$ It requires extensive fine-tuning on QA tasks and has poor generalization ability; $(iii)$ It cannot work for relatively large tables those with over 100+ rows (see more analysis in Section~\ref{sec:sec:pre_limitation}).



There are several challenges in achieving automatic \semiqa. First, understanding the structure of \semitables involves two main problems: $(i)$ how to distinguish header and content cells that could distribute in any areas of the table, which involves semantic understanding (e.g., in the bottom-right table of Figure~\ref{fig:fig1}, the ``A'' cells under ``department'' and ``level'' headers) and cannot be handled by rule-based matching approaches \cite{Burdick2020Table}; 
$(ii)$ how to understand the nested or containment relationships across the header and content cells, where the same question to different cell relationships could lead to different answers. For instance, in the bottom-right table of Figure~\ref{fig:fig1}, when splitting the merged cell of department A, the answer to the question ``what is the second department'' would be department A instead of B (\noindent\textbf{C1}). 
\end{sloppypar}

Second, due to the complexity of \semitable layouts, answering questions over such tables can be challenging, often requiring a variety of analytical ``tricks''. For instance, we may need to apply both left-to-right and top-to-down lookup strategies (e.g., identifying departments in the bottom-right table of Figure~\ref{fig:fig1} by referencing the top-level title header, the left ``Employee Info'' header, and the nested ``Department'' header). Conversely, we may sometimes need to examine the content cells to identify the relevant headers (i.e., bottom-up lookup), which makes the \semiqa procedure even more complicated (\noindent\textbf{C2}).

Third, an effective validation mechanism for \semiqa remains absent, which is especially crucial for resolving issues such as hallucination in \llms~\cite{BeforeGeneration,ChenRTS}. In related tasks like NL2SQL, many methods do not validate the accuracy of generated answers and thus produce the final result in a single shot. Others merely check whether executing the SQL statements yields result tables sufficient to answer the question. However, in \semiqa, the retrieved cells can $(i)$ still involve complex semi-structured layouts (e.g., a single cell may contain multi-row text or even a nested sub-table) and $(ii)$ be derived through multiple lookups, making it difficult for general \llms to verify the accuracy of these retrieved \semitable cells (\textbf{C3}).

To address these challenges, we propose a novel \semiqa framework (\oursys). First, we introduce a graph model (HO-Tree) to represent \semitable layouts, with nodes denoting table headers / content values and edges capturing their hierarchical and containment relationships. This model also includes nine basic tree operations, covering most common QA tasks and addressing the structural complexity challenge (for \textbf{C1}). Second, we present an effective HO-Tree construction strategy: (1) a multi-modal LLM identifies \semitable headers, (2) heuristic rules separate the \semitable into basic table units based on the identified headers, and (3) a depth-first search (DFS) algorithm constructs the HO-Tree. This stage addresses the difficulty of accurately representing the complex implicit relations of \semitables. Third, we propose a question decomposition method with two key techniques: (1) semantic alignment between the input question and the derived operation pipeline, and (2) a column-type-aware tagging approach that annotates discrete, continuous, and unstructured columns (e.g., listing [man, woman] for a \emph{sex} column) to enhance data retrieval accuracy (for \textbf{C2}). Finally, we introduce a two-stage QA verification mechanism to ensure solution stability. The first stage checks constraints (e.g., non-empty, question-related) to validate the generated operations and their execution results. The second stage provides a confidence score by comparing the original questions with those derived from the final answers (for \textbf{C3}). 

We summarize our contributions as follows:

\noindent(1) We present a novel framework that enables effective and robust \semiqa using large language models.


\noindent(2) We provide a tree-based representation method for semi-structured tables (HO-Tree), and design basic tree operations in the model to support common QA tasks. 

\noindent(3) We propose a DFS-based algorithm that combines VLM and heuristic rules to construct HO-Tree from \semitable.

\noindent(4) We design a question decomposition strategy that ensures semantic alignment with the generated operation pipelines, and introduce a column-type-aware tagging strategy to improve lookup accuracy on relatively large \semitables.


\noindent(5) We propose a two-stage QA verification mechanism that conducts constraint examinations and compares the pipelines of the origin question and those derived from the final answer.



\noindent(6) We curate the \emph{SSTQA} dataset, featuring 102 diverse \semitables and 764 representative queries commonly found in real-world scenarios.

\noindent(7) We conduct thorough evaluations to verify \oursys can effectively tackle the structural and semantic complexities of \semitables, resulting in improved QA accuracy and reliability.

\section{Preliminaries}
\label{sec:pre}

In this section, we first explain the typical \semitable layouts, followed by the formalization of the \semiqa task and a discussion of the limitations of existing approaches.

\vspace{-.5em}
\subsection{Semi-Structured Tables}
\label{sec:sec:pre_semitables}

\begin{sloppypar}
{Semi-structured tables can be far more complex than structured ones due to the combination of diverse table layouts. However, compared with other semi-structured data, they still adhere to stricter layout constraints and may suffer information loss when stored in formats like JSON (e.g., splitting merged cells). To preserve their structure, richer formats like HTML~\cite{wikitq} are required. Furthermore, questions over such tables often demand precise layout-aware reasoning (e.g., interpreting layout constraints) and operator grounding (e.g., identifying the intersection of relevant rows and columns).}
\end{sloppypar}


\hi{{Core Elements.}} {In a \semitable $T$, a \emph{table cell} is the atomic unit of a semi-structured table (e.g., the intersection of one row and column). A \emph{table header} consists of one or more rows or columns that label the table body. A \emph{merged cell} spans multiple adjacent rows or columns to convey hierarchical information. {A \emph{subtable} is a semantically and structurally self-contained table embedded within a parent table with at least one level of nested headers, rows, and internal layout.}}


\hi{{Table Layouts.}} {The \semitable $T$ (in standard form and ignoring issues likes data cleaning~\cite{arora2025semi2stru}) can be composed of four typical layouts  (independent with each other):}



\noindent\emph{(L.1)} \bfit{Header-Single-Value.} The simplest layout pairs a header \(H = h_1\) with a single content value \(V = v_1\), arranged either vertically or horizontal:  
\(T_t = \{\;H = h_1,\;V = v_1,\;H\rightarrow V\}\,.\) \emph{For instance, in Table~\ref{tab:layouts}, the atomic header ``Name'' is associated with the single value ``Albert'', demonstrating this minimal \semitable layout.}

\noindent\emph{(L.2)} \bfit{Header-Multiple-Values.} A more prevalent structure in \semitable is an atomic or hierarchical header \(H = h_1\) accompanied by a list of values \(V = [\,v_1,v_2,\dots,v_n\,]\). 
We represent this as \(T_t = \{\;H = h_1,\;V = [\,v_1,v_2,\dots,v_n\,], H\rightarrow V\}\,.\) \emph{For instance, Table~\ref{tab:layouts} presents an example in which the atomic header ``Name'' corresponds to multiple values, specifically the list [\,``Albert'', ``Tim'', ``Jack''\,].}

\noindent\emph{(L.3)} \bfit{Orthogonal-Subtables.} An orthogonal-subtable layout consists of two or more subtables, whose top-level headers appear at the same level, either horizontally or vertically.  {The subtables each contain the same number of rows, forming a parallel structure, while the content values of these subtables are weakly related or unrelated (e.g., personal information for employees in two separate departments).} 
Suppose we have \(n\) such subtables 
\(\,T_1,\, T_2,\, \dots,\, T_n\).  We represent their orthogonal combination as \(T = \Bigl\{\;H {=} [\,H_1, H_2, \dots, H_n\,],\;T {=} [\,T_1, T_2, \dots, T_n\,],\;H_i {\rightarrow} T_i, 1 \leq i \leq n\Bigr\}.\) \emph{For instance, as demonstrated in Table~\ref{tab:layouts}, the three atomic-header-single-value tables are combined in Orthogonal-Table layout.} 

\begin{sloppypar}
\noindent\emph{(L.4)} \bfit{Header‑Orthogonal‑Subtables.} 

{Header-Orthogonal-Subtables consists of an atomic or hierarchical header paired with one or more orthogonal-subtables (\emph{(L.3)}), which means all orthogonal-subtables in \emph{(L.4)} share the same header.} Formally, let the grouped subtables be \(T_1, T_2, \dots, T_n\). We represent this layout as  
\(T_t{=}\Bigl\{
  H {=} H^{\mathsf{p}},
  T {=} [\,T_1,T_2,\dots,T_n\,],
  H^{\mathsf{p}}\!{\rightarrow}\!T_i,\;1\le i\le n
\Bigr\}.\) \emph{For instance, in Table~\ref{tab:layouts}, the header ``Info'' groups two orthogonal subtables (i.e., ``Name'' and ``Age'' subtables) that each follows the Header‑Multiple‑Value layout.}
\end{sloppypar}

A subtable $T_{\text{sub}} \subset T$ follows one or a combination of the above  layouts. Note that: (1) we do not consider irregular layouts, such as content values presented without corresponding headers~\cite{SemiData}; (2) unlike structured tables, common \semitables (e.g., Word tables, transaction records) are relatively small (e.g., tables with over 100 rows are already considered large); (3) we assume the tables are error-free, and issues such as table cleaning (e.g., missing value imputation \cite{alwateer2024missing}) are beyond the scope of this paper. 

\begin{example}
\vspace{-0.7em}
For the bottom-right \semitable in Figure~\ref{fig:fig1}, we can extract two orthogonal subtables sharing a common header ``TD Tech'' ($L.4 \rightarrow L.3$). Within the ``Employee Info'' subtable, we can extract four header-multiple-values subtables ($L.3 {\rightarrow} [L2_1, \dots, L2_4]$). In this way, we can use the formal expression $L.4 {\rightarrow} L.3 {\rightarrow} [L.4 {\rightarrow} \{L.1_1, \dots, L.1_4\}, L.4 {\rightarrow} L.3 {\rightarrow} [L.2[1], \dots, L.2[4]]]$ to recursively traverse the entire structures. Furthermore, the one-to-many relationships between these subtables motivate us to adopt a tree-based strategy to represent \semitables (see Section~\ref{sec:model}).
\vspace{-0.7em}
\end{example}


\begin{table}[t]
    \footnotesize
    \centering
    \setlength{\tabcolsep}{2pt}
    \caption{Semi-Structured Table Layouts -- \textnormal{Cells marked in blue in examples represent table headers.}}
    \vspace{-0.4cm}
    \begin{tabular}{ l l c }
        \hline
        Layout & Formal Representation & Example \\
        \hline
        \makecell{L1: Header- \\ Single-value} & $T_t = \{\;H = h_1,\;V = v_1,\;H\rightarrow V\}\,.$ & 
        \begin{minipage}[c]{0.17\columnwidth}
            \centering
            \raisebox{-.5\height}{\includegraphics[width=\linewidth]{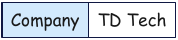}}
        \end{minipage}
        \\
        \hline
        \makecell{L2: Header- \\ Multiple-Values} & \makecell[l]{$T_t = \{\;H = h_1,\;V = [\,v_1,v_2,\dots,v_n\,], $\\$ H\rightarrow V\}\,.$} & 
        \begin{minipage}[c]{0.3\columnwidth}
            \centering
            \raisebox{-.5\height}{\includegraphics[width=\linewidth]{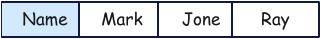}}
        \end{minipage}
        \\
        \hline
        L3: Orthogonal Tables & \makecell[l]{$T = \Bigl\{\;H = [\,H_1, H_2, \dots, H_n\,],\;$\\$T = [\,T_1, T_2, \dots, T_n\,],\;H_i \rightarrow T_i, $\\$ 1 \leq i \leq n\Bigr\}.$} & 
        \begin{minipage}[c]{0.17\columnwidth}
            \centering
            \raisebox{-.5\height}{\includegraphics[width=\linewidth]{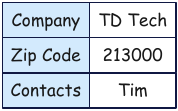}}
        \end{minipage}
        \\
        \hline
    \makecell{L4: Header- \\ Orthogonal-Tables} & \makecell[l]{$T_t=\Bigl\{H = H^{\mathsf{p}},\;T=[\,T_1,T_2,\dots,T_n\,],\;$\\
  $H^{\mathsf{p}}\!\rightarrow\!T_i,\;1\le i\le n \Bigr.$} & 
        \begin{minipage}[c]{0.17\columnwidth}
            \centering
            \raisebox{-.5\height}{\includegraphics[width=\linewidth]{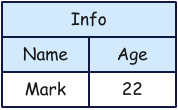}}
        \end{minipage}
        \\
        \hline
    \end{tabular}
    \vspace{-1.em}
    \label{tab:layouts}
\end{table}

\begin{figure}
    \centering 
    \includegraphics[width=1\linewidth]{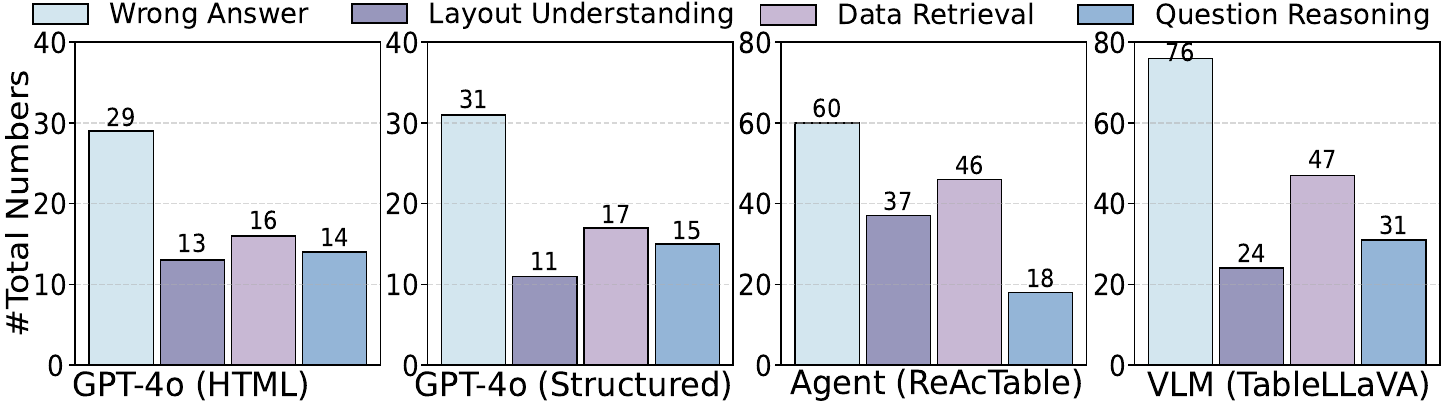}
    \vspace{-2.em}
    \caption{Error Distribution \textnormal{-- {GPT-4o is evaluated on both HTML and structured (JSON) formats; ReAcTable (NL2SQL) converts \semitables into structured representations; while TableLLaVA utilizes LVM to process them as images.}}}
    \label{fig:limitation}
    \vspace{-1.5em}
\end{figure}

\subsection{Semi-Structured Table QA}  
\label{sec:sec:pre_semitable_qa}

Given a \semitable \(T\), 
the QA tasks aim to answer an input question \(Q\) expressed in natural language based on $T$ \cite{schuster2023semqa}.

\begin{definition}[Semi-Structured Table QA]\label{def:rewrite}
Let \(T\) be a semi-structured table with a multi-layered organization, and let \(Q\) be a question that may reference one or multiple subtables in $T$. \Semiqa is defined as a mapping \((T, Q) \;\longmapsto\; A\), where the answer \(A\) is derived by identifying the subtables \(\{T_{\text{sub}} \mid T_{\text{sub}} \subseteq T\}\) relevant to \(Q\).
\end{definition}

\hi{QA Tasks.} Commonly, QA tasks in this problem often require understanding the layouts of target \semitables. Here we showcase three typical QA tasks together with relevant layouts.


\noindent \emph{(1)} \bfit{Numerical Computation.} 
The question $Q2$ in Figure~\ref{fig:intro} requests the age of the oldest employee (\emph{Header-Multiple-Values in \underline{L.2}}). Correctly answering it involves $(i)$ locating the target column ``Age'' within the hierarchical header ``Info'', $(ii)$ extracting all age values from the corresponding records, and $(iii)$ computing the maximum value from the retrieved data. {\it Such QA requires accurate header identification and data retrieval, upon which we can easily apply basic computational functions to get accurate results.}

\noindent \emph{(2)} \bfit{Information Extraction.} The question $Q1$ in Figure~\ref{fig:intro} relies on the ``Employee Info'' layout ({\emph{Header-Orthogonal-Tables in \underline{L.4}}}). To derive the correct answer ``2'', we must extract the records of employees meeting the rating condition and aggregate them. {\it Failure to properly interpret the semantic relationship between the merged cell ``A+'' and its associated employees may lead to a wrong answer.}


\noindent \emph{(3)} \bfit{Summarization.} 
The question $Q3$ in Figure~\ref{fig:intro} requests a summary of the company’s basic information (\emph{Orthogonal-Tables in \underline{L.3}}). Addressing this question involves two critical steps: $(i)$ identifying and extracting the semantically relevant table segments corresponding to the question, and $(ii)$ leveraging the reasoning capabilities of \llms to generate a coherent summary from the retrieved data.
{\it The key challenge lies in robustly interpreting complex table layouts to ensure precise alignment between the question intent and the extracted data to generate accurate summaries.}

\begin{table*}[t]
    \small
    \centering
    \setlength{\tabcolsep}{3pt}
    \caption{Comparison of Relevant Methods for \Semiqa \textnormal{-- More stars indicates better performance.}}
    \vspace{-0.35cm}
    \begin{tabular}{ c c c c c c c l}
        \hline
        Representation & \makecell{Structure \\ Information} & Method & \makecell{Structure \\ Understanding} & \makecell{Data \\ Retrieval} & \makecell{{Supported} \\ {Table Scale}} & \makecell{Answer \\ Accuracy} & Main Challenge \\
        \hline
        HTML/JSON/Spreadsheet & \ding{51} & NL2SQL & - & - & - & - & Fail to operate the table. \\
        HTML & \ding{51} & NL2Code & - & - & - & - & Fail to operate the table. \\
        HTML/JSON & \ding{51} & LLM & \ding{72} & \ding{72} & \ding{72}\ding{72} & \ding{72} & Fail to understand table structure. \\
        JSON/Spreadsheet & \ding{51} & NL2Code & \ding{72} & \ding{72}\ding{72} & \ding{72}\ding{72} & \ding{72}\ding{72} & Fail to understand table structure. \\
        Structured & \ding{55} & NL2SQL / NL2Code & - & \ding{72} & \ding{72}\ding{72}\ding{72} & \ding{72} & Structural information loss.\\
        Structured & \ding{55} & LLM & - & \ding{72} & \ding{72}\ding{72} & \ding{72} & Structural information loss. \\
        Structured & \ding{55} & Agent & - & \ding{72}\ding{72} & \ding{72}\ding{72} & \ding{72} & Structural information loss. \\
        Image & \ding{51} & Multimodal LLM & \ding{72}\ding{72} & \ding{72} & \ding{72} & \ding{72}\ding{72} & Fail to process big tables.\\
        \hline
        HO-Tree & \ding{51} & ST-Raptor & \ding{72}\ding{72}\ding{72} & \ding{72}\ding{72}\ding{72} & \ding{72}\ding{72}\ding{72} & \ding{72}\ding{72}\ding{72} & Modeling and operation. \\
        \hline
    \end{tabular}
    \vspace{-0.35cm}
    \label{tab:methods}
\end{table*}

\vspace{-0.5em}
\subsection{Limitations of Existing Methods}
\label{sec:sec:pre_limitation}

In this section, we discuss the limitations and challenges of existing approaches that could potentially be adapted to \semiqa.
As shown in Table~\ref{tab:methods}, these methods can be categorized based on their table representation strategies (e.g., table serialization~\cite{sui2024table}, HTML/JSON~\cite{herzig2020tapas}) and the question comprehension techniques (e.g.,\cite{qin2022survey}). Figure~\ref{fig:limitation} illustrates the error distribution when applying typical methods to \semiqa.


\hi{(Limitation 1) Poor Table Layout Understanding.} 
The failure in layout understanding indicates that the model incorrectly captures structural evidence, primarily due to the limitations of \semitable representation. Among existing representation methods, structured table representation (i.e., converting \semitables into fully structured formats) leads to major loss of layout information. Alternative approaches (e.g., HTML, JSON, Spreadsheet~\cite{sui2024table}) employ common serialization strategies that partially preserve structural information in textual form. However, due to the inherent one-dimensional nature of these formats, \llms face significant challenges in effectively interpreting complex table layouts.
In contrast, image-based methods exhibit the lowest layout understanding error among wrong answers (31.58\% compared to 35.48\% for GPT-4o with structured input, 44.82\% for GPT-4o with HTML input, and 61.67\% for NL2Code agent), but own relatively low overall accuracy caused by the following two limitations. 
This observation motivates us to \emph{design a proper \semitable representation method that simultaneously stores the structural information and content of \semitables}.

\hi{(Limitation 2) Inaccurate Table Data Retrieval.}
As shown in Figure \ref{fig:limitation}, data retrieval errors constitute a substantial proportion of failures. Specifically, the four methods (i.e., GPT-4o with structured input, GPT-4o with HTML input, NL2Code agent, and vision-language model) exhibit retrieval error rates of 55.17\%, 54.84\%, 76.67\%, and 61.84\%, respectively, which are primarily due to their inability to identify question-relevant tabular data. Among these methods, GPT-4o achieves superior performance, attributable to its advanced contextual comprehension capabilities. In contrast, the agent-based approach, which operates on structured table representations using external tools, incurs significant information loss during structural transformation. We observe that (1) these methods lack robust data retrieval mechanisms, and (2) vanilla \llms struggle to accurately locate target table content, likely due to inherent limitations in semantic understanding. This indicates that \emph{significant improvements remain achievable in accurately locating and retrieving question-relevant content in \semitables}.

\hi{(Limitation 3) Question Comprehension Errors.} 
Question comprehension errors occur when a model misinterprets the semantics of a question and consequently retrieves incorrect answers from \semitables, often due to inadequate integration of table layout and content. VLMs demonstrate the weakest performance, mainly due to their weak understanding of rich-text images. In contrast, both GPT-4o and agent-based methods demonstrate better performance, benefiting from the advanced reasoning capabilities of \llms. However, their remaining errors are predominantly attributable to the failure to align the semantics of the input question with complex semi-structured table layouts, motivating us to \emph{design tailored question decomposition and table semantic alignment techniques for semi-structured tables}.

\vspace{-0.5em}

\section{\oursys Overview} \label{sec:overview}







\begin{figure}
    \centering 
    \includegraphics[width=1\linewidth]{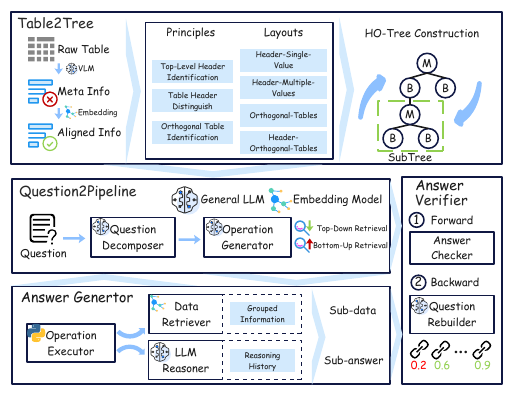}
    \vspace{-2.7em}
    \caption{The \oursys Architecture.}
    \label{fig:arch}
    \vspace{-2.5em}
\end{figure}

\hi{Architecture.} 
Figure~\ref{fig:arch} shows the architecture of \oursys, which consists of four main modules: \bfit{(1) Table2Tree} converts the given semi-structured table into a Hierarchical Orthogonal Tree (HO-Tree), which effectively represents the header / content relationships within the original table (see Section~\ref{sec:sec:tree_construction}). Note that the \emph{Table2Tree} module is utilized only once when handling multiple questions related to the same table. \bfit{(2) Question2Pipeline} transforms complex questions into simpler sub-questions, from which corresponding operation pipeline are  generated for each sub-question (see Section~\ref{sec:sec:operation}). \bfit{(3) AnswerGenerator} then executes these operations to obtain intermediate results or produce the final answer (see Section~\ref{sec:sec:question_decompose}). \bfit{(4) AnswerVerifier} adopts a two-stage validation strategy. In the forward stage, it checks whether the execution results are non-empty and consistent with the question; otherwise, the operation is regenerated or terminated early. In the backward stage, similar questions are generated from the output, and their similarity to the original question is used to score answer's reliability (see Section~\ref{sec:verify}). 

\hi{System Workflow.} When a new semi-structured table and its associated questions arrive, \emph{Table2Tree} first preprocesses the table into an HO-Tree and {serializes the object into a local file}. The \emph{Question2Pipeline} then decomposes each question into subquestions and iteratively interacts with the \emph{AnswerGenerator} to generate answers for each subquestion. The answer to the final subquestion serves as the question answer. The \emph{AnswerVerifier} is involved throughout each step of operation execution, identifying and discarding incorrect intermediate results, based on which \oursys iterates until generating correct answer. 


\begin{figure*}
    \centering \includegraphics[width=1\textwidth]{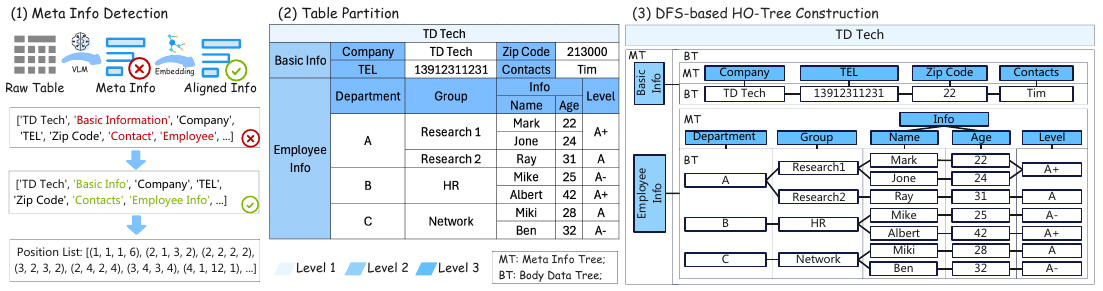}
    \vspace{-2.5em}
    \caption{An example of constructing a three-level nested Hierarchical Orthogonal Tree ($HO\text{-}Tree$) \textnormal{-- Different shades of blue highlight the table's nested levels. For example, in the bottom-right table, the metadata, marked in dark blue, is constructed top-down into a tree of depth two, while the unshaded data section structured into a left-to-right tree of depth five.}}
    \label{fig:tree_construction}
    \vspace{-1.2em}
\end{figure*}

\section{Tree Model for Semi-Structured Table Representation}
\label{sec:model}

As discussed in Section~\ref{sec:sec:pre_semitables}, \semitables consist of complex combinations of basic table layouts (e.g., hierarchical headers with nested subtables). Accurately performing question answering over such tables remains challenging even for advanced \llms like GPT-4o (see Section~\ref{sec:sec:pre_limitation}). Thus, in this section, we study how to effectively represent the layout relationships within \semitables to enable accurate question answering.




\subsection{Hierarchical Orthogonal Tree} \label{sec:sec:ho_tree}


Following the recursive definition in Section~\ref{sec:sec:pre_semitables}, a semi-structured table composed of layouts $L.1$–$L.4$ (Table~\ref{tab:layouts}) can be decomposed into two parts: (1) \emph{Metadata}, which provides high-level semantic abstraction (e.g., headers), and (2) \emph{Data}, which contains the actual content values (e.g., the header ``sex'' associated with [`female', `male', `female']). 
Both metadata and data components may exhibit hierarchical and orthogonal structures. Given this inherent one-to-many organization, we model them using trees: one for metadata and one for data. In each, nodes store metadata or content values, while edges encode containment or orthogonal relationships.

\begin{definition}[Hierarchical Orthogonal Tree ($HO\text{-}Tree$)\label{def:tree}] Given a \semitable $T$, we model $T$ into HO-Tree that links the metadata with corresponding content values by pointing the leaf node in the $MTree$ to each level of $BTree$, representing the association between metadata and the corresponding table column:\\
(1) {\bf Meta Tree ($MTree$).}  It represents the structural information and content of the table's metadata. Each path from the root to a leaf corresponds to an abstract description of a specific column. Nodes in \(MTree\) are denoted as \(MNode\);\\
(2) {\bf Body Tree ($BTree$).} It represents the structural information and content of the table body. Each node corresponds to a single cell value, each path from the root to a leaf represents a row, and each level of the tree corresponds to a column—aligned with a path in \(MTree\). Nodes in \(BTree\) are denoted as \(BNode\).\\
A single \(HO\text{-}Tree\) is represented as \(T = \{MTree = M, BTree = B, M \rightarrow B\}\).  
Within a \semitable, a collection of such \(HO\text{-}Trees\) may exist, each comprising an \(MTree\) and a \(BTree\). Moreover, hierarchical containment may exist among these trees, where one \(HO\text{-}Tree\) can serve as the value of a node within another \(BTree\).
\end{definition}

\begin{example}
The right part of Figure~\ref{fig:tree_construction} illustrates an example of a HO-Tree, where ``TD Tech'' serves as the top-level cell. This cell is stored in a single-node \(MTree\), which points to the remaining structure stored in a \(HO\text{-}Tree\). The subsequent \(MTree\) contains two \(MNode\) instances (i.e., ``Basic Info'' and ``Employee Info''), each referencing a distinct \(BNode\), with each \(BNode\) storing a sub $HO\text{-}Tree$.
\end{example}

\begin{table}[t]
    \small
    \centering
    \setlength{\tabcolsep}{2pt}
    \caption{Table of Atomic Operations.}
    \vspace{-0.4cm}
    \begin{tabular}{ l l l }
        \hline
        Operation & Formal Representation & Description \\
        \hline
        Children & $CHL(V)$ & Get children based on the given value.\\
        Father & $FAT(V)$ & Get Father based on the given value.\\
        Value & $EXT(V_1,V_2)$ & Get the cross of two values.\\
        Condition & $Cond(D,Func)$ & Filter values based on the function.\\
        Calculation & $Math(D,Func)$ & Calculate result based on the function.\\
        Compare & $Cmp(D_1,D_2,Func)$ & Compare values based on the function.\\
        Execute & $Foreach(D,Func)$ & Apply function to the given values.\\
        Align & $Align(P,HO\text{-}Tree)$ & Operation-Table alignment.\\
        Reason & $Rea(Q,D)$ & Reasoning based on \llms.\\
        \hline
    \end{tabular}
    \vspace{-0.1cm}
    \label{tab:operation}
\end{table}

\subsection{HO-Tree Construction} \label{sec:sec:tree_construction}

Based on the definition, constructing an HO-Tree from a given \semitable requires precisely identifying (1) meta-information (table headers), (2) content values, as well as (3) the relationships between subtables. Two main challenges remain. First, headers and content cells can appear in arbitrary positions, making it difficult to distinguish between them. Second, understanding the nested or containment relationships among these cells is non-trivial, especially given the flexible and irregular structure of \semitables. Thus, next we introduce the detailed steps in HO-Tree construction.

\vspace{-1.em}
\begin{algorithm}[htbp]
    \LinesNumbered
    \KwIn{A Semi-Structured Table $T$}
    \KwOut{Extracted HO-Tree $HOTree$}
    \BlankLine
    \caption{HO-Tree Construction (HOTC)} \label{algo:tree_construction}
    $MetaInfo \gets MetaInfoDetect(T)$\;
    $T_{list} \gets TablePart(T, MetaInfo)$\; 
    $HOTree_{list} \gets []$\;
    \For{$T_{sub}$ in $T_{list}$}{
        \Switch{$type(T_{sub})$} {
            \uCase{$L_1,L_2,L_4$} {
                $HOTree_{list}.push\_back(ConsTree(T_{sub}))$\;
            }
            \uCase{$L_3$} {
                $HOTree_{list}.push\_back(HOTC(T_{sub}))$\;
            }
        }
    }
    \Return{$ConsTree(HOTree_{list})$}\;
\end{algorithm}
\vspace{-1.em}

\subsubsection{Meta Information Detection}
\label{sec:sec:meta_info_detect}

Unlike structured tables with fixed schemas, \semitables often exhibit complex and irregular nesting, posing challenges for meta-information extraction. Rule-based approaches fail to capture such nuances, especially when similar structural patterns represent different semantics. Serializing tables (e.g., as images or structured formats) allows prompting models for metadata extraction or format generation. However, LLMs trained on 1D text struggle with 2D tables~\cite{Table-GPT}, often exhibiting issues like hallucination and the ``Lost in the Middle'' effect~\cite{lostinthemiddle}, undermining consistency and reliability.

To overcome these limitations, we adopt a hybrid method that combines rule-based matching with LLM-based reasoning. 
\revisiona{As shown in Figure~\ref{fig:tree_construction}, given a semi-structured table in Excel format, we first convert it to HTML, render it using a headless browser, and capture a high-resolution screenshot as input image for the VLM. Then, we prompt the VLM to output all possible keys that would be present in a JSON-formatted representation of the table as the candidate meta-information cell values.}
Subsequently, we calculate similarity scores between candidates and all table cells using embedding-based similarity metrics. Cells exceeding a predefined threshold are identified as meta-information, and their positions guide the subsequent table partitioning process.

\subsubsection{Table Partition Principles} 
\label{sec:sec:table_process_principle}

We introduce three principles to guide the HO-Tree construction process by interpreting different layouts in \semitables:

\hi{(Principle 1) Top-Level Header Identification.} If a merged cell spans an entire row or column, it is treated as a header in a Header-Orthogonal-Tables layout ($L.4$), and adjacent cells (below or to the right) are interpreted as a subtable.


\hi{(Principle 2) Header-Content Differentiation.}
When both top-aligned and left-aligned headers are present, the one with more cells is selected to construct the $MTree$, while the other is integrated into the $BTree$.

\hi{(Principle 3) Orthogonal Table Identification.}
When Orthogonal-Tables ($L.3$) are detected, we segment them and process each subtable recursively and sequentially.


As illustrated in Figure~\ref{fig:tree_construction}, applying these principles enables table partitioning based on meta-information locations and the recursive construction of HO-Trees for \semitables.

\vspace{-0.5em}
\subsubsection{DFS-based Tree Model Construction} 
\label{sec:sec:dfs_tree_construction}

An HO-Tree is then constructed for each subtable according to its identified layout type: for layouts $L.1$ and $L.2$, the HO-Tree is built directly; otherwise, a recursive DFS is applied. For example, in Figure~\ref{fig:tree_construction}, the vertically structured \semitables demonstrate a top-down alignment of metadata and content.

Based on the detected metadata, the table is partitioned into a list of subtables following predefined principles. An HO-Tree is then constructed for each subtable according to its identified layout type: for layouts $L.1$ and $L.2$, the HO-Tree is built directly; otherwise, a recursive DFS is applied. Take the vertically structured \semitables where metadata and content are aligned top-down in Figure~\ref{fig:tree_construction} as an example: 
(1) In the $MTree$, each root-to-leaf path represents a column, and the tree grows vertically to reflect vertical relationships among metadata.
(2) In the $BTree$, each path represents a row, with horizontally aligned attributes, so the tree grows horizontally. 
Each leaf node in the $MTree$ points to a corresponding level in the $BTree$, linking metadata to the associated content column.

During DFS backtracking, we model Orthogonal-Tables and Header-Orthogonal-Tables layouts (i.e., $L.3$ and $L.4$). In such cases, a node in the $BTree$ may recursively contain another HO-Tree as its value.





\revisionb{Algorithm~\ref{algo:tree_construction} outlines the HO-Tree construction process. 
VLMs identify table headers via the $MetaInfoDetect$ function (Section~\ref{sec:sec:meta_info_detect}). 
Using the extracted metadata and predefined principles, the table is partitioned into subtables through the $TablePart$ function (Section~\ref{sec:sec:table_process_principle}).
To address structural complexity, each subtable is transformed into an HO-Tree via the $ConsTree$ function (Section~\ref{sec:sec:dfs_tree_construction}) and recursively merged using depth-first search to reconstruct the full structure of the original \semitable.}

\begin{example} 
\vspace{-0.5em}
\revisionc{Figure~\ref{fig:tree_construction} illustrates the construction of a three-level nested HO-Tree. The ``Basic Info'' subtable is initially misidentified by the VLM as ``Basic Information'' and corrected by alignment. Metadata is used to identify the $L.4 \rightarrow L.3$ layout, guiding the partition of the ``Basic Info'' subtable into $MNode$ and a sub-$HO\text{-}Tree$ (constructing $L.4$). Finally, the leaf nodes of the $MTree$ in subtree link to corresponding levels of the $BTree$ (constructing $L.3$).}
\vspace{-0.5em}
\end{example}


\begin{figure*}
    \vspace{-1em}
    \centering \includegraphics[width=1\textwidth]{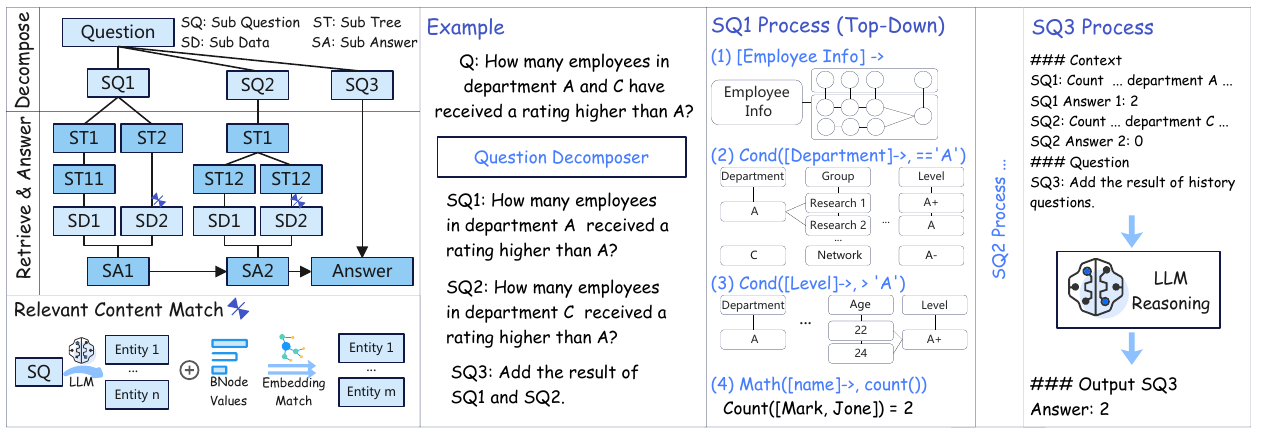}
    \vspace{-2.5em}
    \caption{Question Decomposition and Pipeline Generation.}
    \vspace{-1.5em}
    \label{fig:operation_execution}
\end{figure*}


\vspace{-1.25em}
\section{Pipeline-based Question Answering} 
\label{sec:pipeline_qa}


With the HO-Tree model in place, it poses three key challenges in question answering over \semitables. First, such questions often require multi-hop reasoning rather than one-shot retrieval, and unlike NL2SQL tasks, lack a standardized operation set for pipeline construction. Second, answering necessitates hybrid traversal strategies (e.g., left-to-right, top-down, bottom-up) to locate relevant content. Third, the large number of cells in some \semitables complicates the precise identification of question-relevant information.

To address these challenges, we propose a pipeline-based QA strategy centered on a set of tree-specific operations that cover most QA scenarios: (1) a question decomposition method for complex multi-hop queries, (2) an operation generation mechanism with parameter-content alignment, and (3) a column-type-aware tagging mechanism that annotates columns based on data characteristics, enabling efficient and accurate retrieval from large, complex tables.

\subsection{Basic Operations over HO-Tree} \label{sec:sec:operation}


We design a suite of atomic operations to enable structured and interpretable QA over the $HO\text{-}Tree$. These operations support both precise tree traversal and auxiliary tasks correspond to common \semitable sub-tasks. The operations fall into four categories: \emph{(1) Data Retrieval Operations}, which retrieve relevant values from the tree; \emph{(2) Data Manipulation Operations}, which process or transform retrieved data; \emph{(3) Alignment Operations}, which align operation parameters with table content; \emph{(4) Semantic Reasoning Operations}, which invoke \llms for contextual inference.


For any user question, we first decompose it into one or more sub-questions, each resolved through a sequence of operations to retrieve relevant information or derive the answer. Ideally, the generated operation pipeline includes: (1) retrieval to collect non-redundant data, (2) manipulation to structure the data into a model comprehensible form, (3) alignment to ensure question-content alignment, and (4) reasoning to produce the final answer. In worst-case scenarios, the model may retrieve nearly the entire table and rely heavily on reasoning, highlighting the need for fine-grained, modular operation design.

\hi{Data Retrieval Operation.} 
These operations are responsible for extracting relevant table content from the $HO\text{-}Tree$ based on arguments derived from the user question.

\noindent$\bullet$ \bfit{Children Retrieval ($CHL(V)$)} 
This operation retrieves all successor nodes of any node whose value matches $V$. If multiple such nodes exist, each set of successors is returned separately. 
Use the $HO\text{-}Tree$ in Figure \ref{fig:tree_construction} as an example, $CHL(Basic\ Info)$ returns the $HO\text{-}Tree$ containing the company information.


\noindent$\bullet$ \bfit{Father Retrieval ($FAT(V)$)} 
This operation is used to obtain the set of ancestor nodes of a given tree node. 
For instance, in Figure \ref{fig:tree_construction}, $FAT(Department)$ would return the $HO\text{-}Tree$ containing the employee information.


\noindent$\bullet$ \bfit{Value Retrieval ($EXT(V_1, V_2)$)} 
This operation is used to find nodes in a certain layer of $BTree$ pointed to by a $MTree$ leaf node, while giving them a common ancestor $BNode$ as a filtering criterion. Specifically, one of $V_1$ and $V_2$ needs to be a $MNode$ value, and the other needs to be a $BNode$ value. Assume that $V_1$ is the $MNode$ value and $V_2$ is the $BNode$ value, the operation returns a set of node values that satisfy the following: (1) The $BNode$ is in the $MTree$ column $V_1$ (2) $BNode$ with value $V_2$ is an ancestor of the $BNode$ in $MTree$ column. 

\hi{Data Manipulation Operation} is used to perform specific operations on the data, including filtering based on conditions, performing calculations, and making comparisons.

\noindent$\bullet$ \bfit{Condition ($Cond(D, Func)$)} filters a data set $D$ using a predicate function $Func$ and returns the filtered values or a new $HO\text{-}Tree$. 
For instance, $Condition(EXT(Lily, Grade), def(x):\ return\ x\ <\ 60)$ retrieves Lily's grades that are less than 60.

\begin{sloppypar}
\noindent$\bullet$ \bfit{Calculation ($Math(D, Func)$)} applies a numerical computation function $Func$ over a data set $D$.
For instance, $Math(CHL(Price) , def(x):\ return\ sum(x))$ returns the sum of the column ``Price''.

\noindent$\bullet$ \bfit{Compare ($Cmp(D_1, D_2, Func)$)} compares two data sets $D_1$ and $D_2$ using a function $Func$, and returns the boolean result.
For instance, $Compare(EXT(Lily,Grade), EXT(Cindy, Grade), def(x1,x2): return\ x1>x2$ returns the truth value of the statement ``Lily's grade is greater than Cindy's''.
\end{sloppypar}

\noindent$\bullet$ \bfit{Execute ($Foreach(D, Func)$)} applies a function $Func$ to each element in data set $D$ and returns the resulting set. 
For instance, $Foreach(EXT(Lily, Grade), def(x)\ return\ x-10)$ retrieves all Lily's grades and returns the values after minus ten.

\hi{Alignment Operation} aims to ensure consistency between operation parameters and the content of the $HO\text{-}Tree$.

\noindent$\bullet$ \bfit{Align ($Align(P, HO\text{-}Tree)$)}
This operation aligns the parameters $P$ in a given operation with the nodes in the $HO\text{-}Tree$ using an embedding-based similarity model~\cite{MultilingualE5}. The process involves computing embeddings for the parameters and table content:


\begin{gather}
E_a = Embed(a_1,a_2,\dots,a_n)  \\
E_c = Embed(c_1,c_2,\dots,c_m)  \\
SimMatrix = CosSim(E_a, E_c)
\end{gather}
where $a_i$ represents the $i$-th parameter and $c_j$ denotes the content of the $j$-th node in the tree, with $m$ representing the total number of nodes. Cosine similarity ($CosSim$) is used to identify the most semantically aligned table node for each parameter, which is then used for downstream operations.

\begin{example}
\vspace{-0.5em}
\revisionc{For the operation $CHL(Identification)$, if the meta-information of the table is [`ID', `Name', `Age'], the alignment operation $Align(Identification, HO\-Tree)$ attempts to output ``ID'' to align the operation parameter to the table content.}
\vspace{-1.em}
\end{example}


\hi{Semantic Reasoning Operation} leverages LLMs for high-level reasoning over retrieved data.

\noindent$\bullet$ \bfit{Reason ($Rea(Q, D)$)} 
This operation takes the original question $Q$ and the data $D$ obtained from prior operations, and uses an \llm to generate the final answer. In cases where the question is decomposed into multiple sub-questions, semantic reasoning can also be used to aggregate intermediate answers into a final response.

\subsection{Question Decomposition and Pipeline Generation} \label{sec:sec:question_decompose}

Figure~\ref{fig:operation_execution} illustrates the overall process of question decomposition, operation generation, and step-by-step data retrieval via both top-down and bottom-up strategies.

\bfit{Step 1: Question Decomposition.} As shown in Figure~\ref{fig:operation_execution}, upon receiving a user question, we first utilize the semantic understanding capabilities of an \llm to decompose complex multi-hop questions into multiple simpler, single-step sub-questions. \revisiona{Specifically, we prompt the LLM with the input question, sampled table content as semantic supplements, and example decomposition cases that are dynamically retrieved based on question similarity to generate sub-questions.} These sub-questions are often interdependent: some can be directly resolved using the $HO\text{-}Tree$, while others rely on intermediate results produced by earlier sub-questions. For example, a question such as ``What is the total salary for 2021 and 2022?'' can be decomposed into three sub-questions: (1) retrieve the 2021 salary, (2) retrieve the 2022 salary, and (3) compute the sum of the two. \revisiona{For table operation generation, we prompt the LLMs with a predefined set of operations and illustrative examples.}


\bfit{Step 2: Relevant Data Retrieval.} \revisionc{Each sub-question is then answered independently through a combination of top-down and bottom-up retrieval. 
\oursys always starts with top-down retrieval and switches to bottom-up retrieval when the former fails.}
The left section of Figure~\ref{fig:operation_execution} shows this iterative process. For each sub-question, the \llm first generates an operation statement based on the sub-question content and the meta-information extracted from the $HO\text{-}Tree$. Executing this operation yields intermediate results (i.e., sub-trees), which are then used to inform subsequent operation generation.

\revisionc{We employ the \texttt{Align} operation for two purposes: (1) Operation-Table Alignment, applied upon each operation's generation to align sub-questions with the corresponding table content, and (2) Relevant Content Match, invoked when the number of data nodes is large, to extract key entities from the question and constrain the search space within the HO-Tree.}

\bfit{Step 3: Answer Generation}
Once the relevant sub-data is retrieved, each sub-question is resolved via the \texttt{Reason} operation, and all sub-answers are aggregated to produce the final answer.

\subsection{Relatively Large Table QA Enhancement} \label{sec:sec:grouping}

\begin{figure}
    \centering 
    \includegraphics[width=1\linewidth]{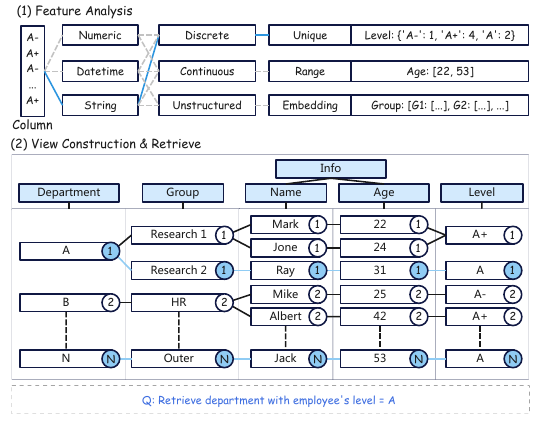}
    \vspace{-2.5em}
    \caption{Characteristic-based Grouping Mechanism}
    \label{fig:grouping}
    \vspace{-2.em}
\end{figure}


The abundance of available \emph{BNodes} poses challenges in selecting accurate operation parameters. To address this, we introduce a data grouping strategy that leverages inherent structural and semantic features, combining top-down grouping based on data characteristics with parallel grouping via tree structure traversal.


\hi{1. Characteristic-based Grouping.} 
 \revisionc{The top-down grouping process is performed during HO-Tree construction, clustering data based on column characteristics.} 
As shown in Figure~\ref{fig:grouping}, column values are classified by data type (e.g., Numeric, Datetime, Unstructured String) and then further categorized as follows: \bfit{Discrete} — columns with a limited set of values (e.g., grades [`A', `B', `C', `D'] or binary options [Yes, No]); \bfit{Continuous} — columns with numeric values spanning a range (e.g., height or temperature); and \bfit{Unstructured} — columns with free-form text (e.g., comments or descriptions).
 \revisiona{A rule-based classifier performs this categorization to reduce complexity in tables with numerous similar cells. With clearly defined rules, it achieves near-perfect accuracy.}


Each category is then grouped using tailored strategies: (1) \bfit{Discrete} values are clustered by exact matches (e.g., students with Grade A); (2) \bfit{Continuous} values are sorted and partitioned into fixed intervals to preserve numeric order (e.g., stress levels by range); and (3) \bfit{Unstructured} values are grouped via embedding-based clustering to capture semantic similarity.


As shown in Figure~\ref{fig:grouping}, a column with values like Grade (A+, A, A\text{-}, ...) is classified as \bfit{Discrete}, allowing identical values to be grouped. Queries such as ``fetch all students with Grade A'' can then be resolved efficiently within the corresponding group.

\hi{2. Group-based Data Retrieval.} 
The parallel-direction grouping leverages structural relationships within the $HO\text{-}Tree$. After locating the target node(s) based on content, the tree is traversed to retrieve related information by: (1) searching \emph{upwards} for ancestor nodes and (2) searching \emph{downwards} for descendant nodes.

This bi-directional traversal allows for the reconstruction of a minimal sub-tree that contextualizes the target node, revealing the complete information associated with the same entity (e.g., retrieving all attributes of a specific student or transaction).

\section{Two-Stage QA Verification}
\label{sec:verify}

Robust validation mechanisms are essential for the reliability of \semiqa, particularly given that existing solutions (e.g., NL2SQL~\cite{PETSQL,OpenSearchSQL}) often lack comprehensive answer verification. The main challenge arises from the fact that, different from structured table QA, the retrieved result cells in \semiqa often exhibit complex layouts and may be derived through multi-step lookup processes, which is tricky for general \llms to verify. To address this problem, we propose a two-stage verification framework that integrates both forward and backward validation to enhance the robustness and trustworthiness of the final answers.


\hi{Forward Verification.} The first stage focuses on validating the correctness of intermediate operations and execution results during the QA process. Specifically, at each step of generating and executing operations, we verify whether the parameters produced by \llm are consistent with the actual contents of the table (e.g., when asking for company information and the given data only contain employee information, we stop the process). This involves directly matching generated parameters against table cells to ensure semantic and syntactic alignment.

After executing each operation over the HO-Tree, the general \llm evaluates whether the resulting data sufficiently answers the corresponding sub-question. If the result is found to be inadequate, a new operation statement is generated to continue the reasoning process. Notably, real-world queries often lack sufficient information to be answered directly from the table, which may cause the model to hallucinate incorrect answers. To mitigate this issue, the forward verification mechanism allows the model to halt the pipeline and return an ``unanswerable'' signal when it detects that the answer cannot be reliably inferred. Furthermore, when operation statements are found to misalign with the table content, the system triggers a regeneration process to refine and correct the statements, thereby improving both the data retrieval accuracy and efficiency.


\hi{Backward Verification.} 
In the second stage, we evaluate the correctness of the generated operation pipeline by verifying it against alternative reasoning paths. Notably, different questions over the same \semitable may yield identical answers via distinct yet similar pipelines (e.g., ``Who is the highest-paid employee?'' vs. ``Who leads the technical department?''). 
\revisionc{Leveraging this property, we perform backward verification by generating alternative questions with the same answer, deriving corresponding pipelines, and measuring their similarity to the original. The average similarity serves as an implicit indicator of pipeline and answer reliability.}
\revisionc{Specifically, we employ few-shot learning, retrieving question-similar examples to guide the generation of alternative questions to solve the problem that questions with the same answers may show tremendous difference in their operation pipeline.}



\vspace{-1.5em}
\section{Experiments} \label{sec:experiment}

\subsection{Experiment Setup}    \label{sec:sec:experiment_setup}

\hi{LLMs.} We use InterVL2.5 26B~\cite{InterVL} as the vision language model, Deepseek-V3~\cite{DeepSeekV3} as the general LLM, and Multilingual-E5-Large~\cite{MultilingualE5} as the semantic embedding model. 

\begin{sloppypar}
\hi{Evaluated Methods.} The evaluated methods include: (1) NL2SQL. OpenSearch-SQL~\cite{OpenSearchSQL} employs a dynamic few-shot learning strategy (Query-CoT-SQL) and introduces an SQL-Like intermediate language to optimize reasoning chains. (2) Foundation Model. GPT-4o~\cite{GPT4}, the cutting-edge \llm developed by OpenAI, and DeepSeekV3~\cite{DeepSeekV3}, a strong Mixture-of-Experts LLM developed by DeepSeek-AI is evaluated. (3) Fine-tuning based Methods. TableLLaMA~\cite{TableLLaMA} is a fine-tuned version of LLaMA-2 7B tailored for tabular data processing across eight table-specific tasks. The model handles various table types, including Wikipedia tables and spreadsheets. TableLLM~\cite{TableLLM} is a 13B-parameter \llm designed for tabular data manipulation tasks, which is fine-tuned on a diverse mix of table-centric datasets in processing document tables and spreadsheets, making it well-suited for real-world office scenarios. (4) Agent based Methods. ReAcTable~\cite{ReAcTable} is an agent-driven approach that integrates reasoning and action-based decision-making for table question answering. It iteratively generates operations, updates the table, and constructs a reasoning chain as a proxy for intermediate thought processes through prompting LLMs and in-context learning. TAT-LLM~\cite{TATLLM} extracts relevant segments from the context, generates logical rules or equations, and then applies these rules or executes the equations to derive the final answer through LLM prompting. (5) VLM based Methods. TableLLaVA~\cite{TableLLaVA} extends the training of LLaVA-7B/13B on 150K table recognition samples, allowing the model to align table structures and elements with textual modality. We choose the 7B version in our experiment. mPLUG-DocOwl1.5~\cite{DocOwl1.5} is a fine-tuned VLM with 8B parameters. It incorporates a spatial-aware vision-to-text module designed to represent high-resolution, text-rich images while preserving structural information and reducing the length of visual features.

\hi{Input Formats.}
\revisionb{NL2SQL and agent-based methods take structured tables as input, typically stored in databases or CSV files. Fine-tuning-based approaches operate on structured tables in Markdown format, while VLM-based methods accept table images as input. Foundation models generally utilize the HTML representation of tables. \oursys currently supports Excel as the input format and is compatible with all lossless table representations like HTML.}

\hi{Benchmarks.} \revisionb{We evaluate nine baselines and \oursys on three benchmarks: $(i)$ WikiTQ, featuring Wikipedia tables with complex natural language questions, $(ii)$ TempTabQA, targeting on temporal question answering over semi-structured tables, and $(iii)$ SSTQA, our proposed dataset detailed in Section~\ref{sec:sec:experiment_data}. Since the table formats in other datasets do not fully adhere to the definition of semi-structured tables, we select a subset of tables that meet the criteria and denote them as WikiTQ-ST and TempTabQA-ST, respectively.}

\end{sloppypar}

\begin{table}[!t]
    \small
    \centering
    \setlength{\tabcolsep}{3pt}
    \caption{\revisionc{Characteristics of SSTQA Benchmark.}}
    \vspace{-1.5em}
    \begin{tabular}{ c c c c c c c}
        \hline
        Dataset & \makecell{Nesting Depth} & \makecell{Merge  Ratio} & \makecell{Cell\\Count} & \makecell{Avg. length \\ of Contents}\\
        \hline
        WikiTQ & 1.2970 & 0.0091 & 178.3564 & 1.9568 \\
        TEMPTABQA & 2.0000 & 0.1780 & 44.8350 & 3.6696 \\
        INFOTABS & 2.0000 & 0.0548 & 23.6683 & 2.0769 \\
        SSTQA & 2.5196 & 0.0544 & 147.4608 & 2.7287 \\
        \hline
    \end{tabular}
    \vspace{-3.em}
    \label{tab:dataset_comp}
\end{table}

\vspace{-0.5em}
\subsection{SSTQA Benchmark}  \label{sec:sec:experiment_data}
\revisionb{Existing semi-structured table datasets face two key limitations: (1) they consist of small, structurally simple tables that fail to evaluate a model’s capacity to comprehend complex \semitables; and (2) their queries are misaligned with practical applications, limiting real-world utility.}
\revisionb{For example, although WikiTQ~\cite{wikitq} includes a large number of semi-structured tables from Wikipedia, these tables typically exhibit simple layouts with merged cells and are converted into structured formats as part of the benchmark preprocessing. Meanwhile, TempTabQA~\cite{temptabqa} consists solely of shallowly nested tables with fewer than five columns, lacking the structural complexity commonly observed in real-world datasets.}  

To fill the gap, we introduce SSTQA, a dataset specifically designed to evaluate a model’s ability to conduct \textbf{S}emi-\textbf{S}tructured \textbf{T}able \textbf{Q}uestion \textbf{A}nswering task in real-world scenarios. \revisionb{As shown in Table~\ref{tab:dataset_comp}, WikiTQ has the highest cell count but the shallowest nesting depth. In contrast, SSTQA exhibits the deepest nesting and the relatively large table size among the existing semi-structured table benchmarks.}

\hi{Data Collection.}
\revisiona{The 102 tables in SSTQA are carefully curated from over 2031 real-world tables coverage across 19 representative real scenarios (e.g., administrative and financial management) by considering tables featuring semi-structured formats, such as nested cells, multi-row/column headers, irregular layouts, which ensures the representativeness both in structure and information.}

\revisiona{For question-answer pair generation, we employ a two-stage approach. First, we augment the question set by extracting information from tables as answers, then generating corresponding questions to enhance QA pair alignment. Second, we sample question templates and prompt a \llm to generate open-ended question-answer pairs based on the table and template.
To ensure data quality, we implement a two-step verification process. Initially, a \llm validates the alignment between tables, queries, and answers. This is followed by manual inspection to verify answer correctness by 11 professional annotators, which results in a high-quality dataset of 764 meticulously curated table-based QA pairs.}



\hi{Table Complexity.} \revisiona{We categorize table difficulty based on a weighted combination of three key features: (\romannumeral1) nesting depth (0.5), (\romannumeral2) structural irregularity, including the number of header rows and column spans (0.3), and (\romannumeral3) average cell content length (0.2). After z-score normalization and feature aggregation, SSTQA tables are grouped into 59 simple, 33 medium, and 10 hard instances.}

\hi{Table + QA Complexity.} \revisiona{Table-question difficulty varies with the combination of table structure and query complexity. Accordingly, we categorize Table+QA tasks into three levels: (\romannumeral1) simple, where answers can be directly retrieved from the table; (\romannumeral2) medium, requiring logical inference or conditional operations; and (\romannumeral3) hard, where answers are not explicitly present and demand semantic reasoning. We obtain 299 simple, 284 medium, and 178 hard cases. The corresponding experimental results are presented in section~\ref{sec:sec:experiment_analysis}.}


\begin{table}[!t]
    \small
    \centering
    \vspace{-0.7cm}
    \setlength{\tabcolsep}{3pt}
    \caption{\revisionb{Overall Performance Comparison.}}
    \vspace{-0.2cm}
    \begin{tabular}{ l c c c c c c}
        \hline


        

        \multirow{2}{*}{Methods} & WikiTQ-ST & TempTabQA-ST & \multicolumn{2}{c}{SSTQA} \\

        & Acc & Acc & ROUGE-L & Acc \\
        OpenSearch-SQL~\cite{OpenSearchSQL} & 38.89 & 4.76 & 23.87 & 24.00 \\
        TableLLaMA~\cite{TableLLaMA} & 35.01 & 32.70 & 26.71 & 40.39 \\
        TableLLM~\cite{TableLLM} & 62.40 & 9.13 & 2.93 & 7.84 \\
        ReAcTable~\cite{ReAcTable} & 68.00 & 35.88 & 7.49 & 37.24 \\
        TAT-LLM~\cite{TATLLM} & 23.21 & 61.86 & 19.26 & 39.78 \\
        
        TableLLaVA~\cite{TableLLaVA} & 20.41 & 6.91 & 5.92 & 9.52 \\
        mplug-DocOwl1.5~\cite{DocOwl1.5} & 39.8 & 39.80 & 28.43 & 29.65 \\
        GPT-4o~\cite{GPT4} & 60.71 & 74.83 & 43.86  & 66.45\\
        DeepSeekV3~\cite{DeepSeekV3} & 69.64 & 63.81 & 46.17& 63.22  \\
        \textbf{ST-Raptor (Ours)} & \textbf{71.17} & \textbf{77.59} & \textbf{52.19} & \textbf{72.39} \\

        \hline
    \end{tabular}
    \vspace{-0.6cm}
    \label{tab:experiment}
\end{table}

\hi{Evaluation Metrics.} We adopt two primary evaluation metrics: Answer Accuracy (Acc), following prior work~\cite{TableLLaMA,Table-GPT}, and ROUGE-L to accommodate summarization-style questions in SSTQA. To address the limitations of exact string matching, we further employ general-purpose LLMs to compare model predictions and ground-truth answers, enabling a more nuanced evaluation.








\begin{figure}
    \centering \includegraphics[width=1\linewidth]{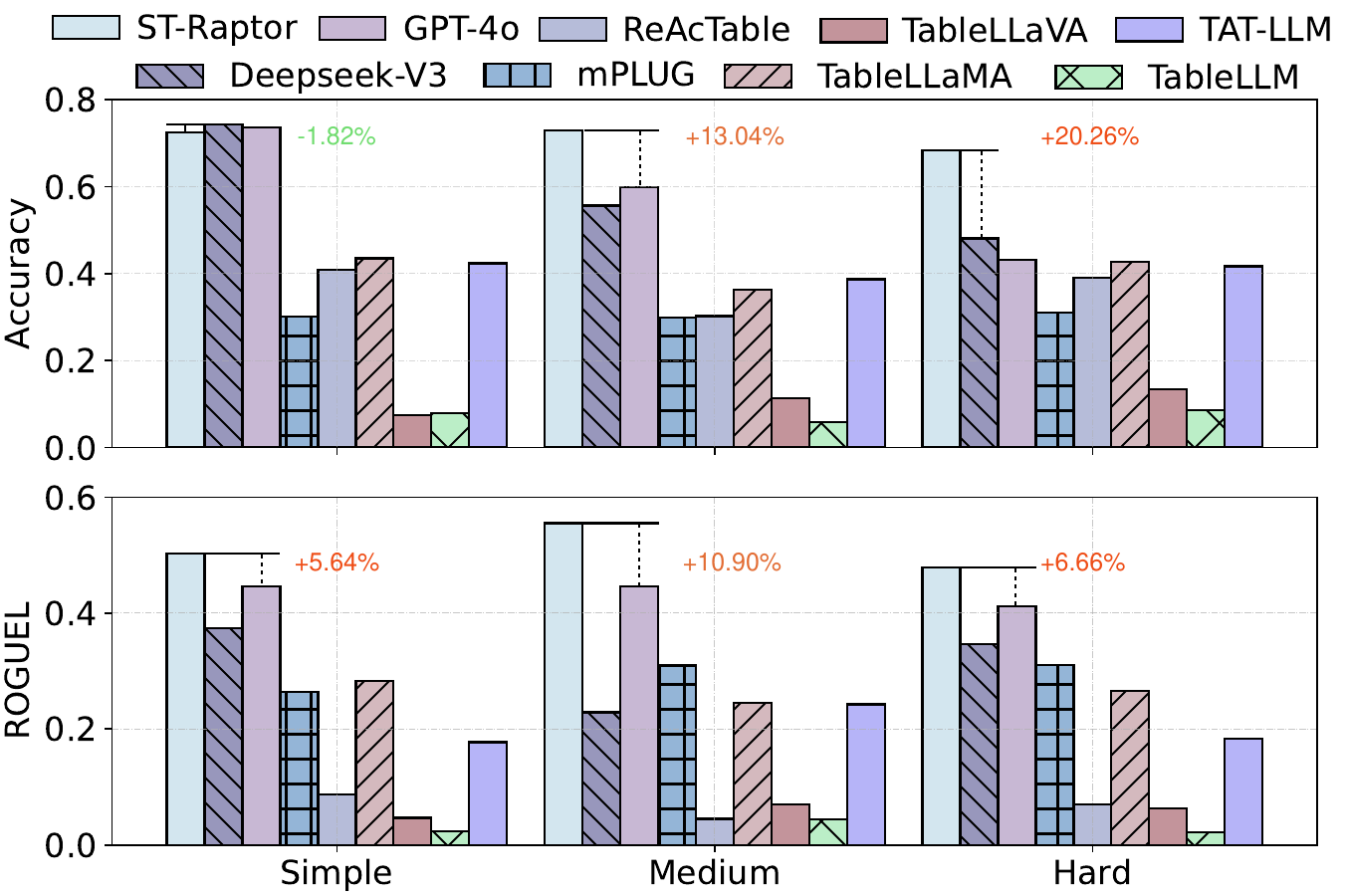}
    \vspace{-2.5em}
    \caption{Evaluation Results under Different Table Difficulty.}
    \label{fig:table_difficulty}
    \vspace{-2.5em}
\end{figure}

\subsection{Overall Performance Comparison}
\label{sec:sec:experiment_analysis}

We conduct a comprehensive evaluation of \oursys against nine state-of-the-art table question answering (QA) methods, spanning five technical paradigms, i.e., NL2SQL, fine-tuning based methods, agent based methods, VLM based methods, and foundation \llms.
The general accuracy of these methods across three \semiqa benchmarks is shown in Table~\ref{tab:experiment}. Then, we further classify the difficulty of \semitables into Simple, Medium, and Hard tiers, and visualize the accuracy variation upon different table difficulties, which is shown in Figure~\ref{fig:table_difficulty}.

s\hi{Overall Accuracy.}
Experiments show that \oursys consistently outperforms all nine baselines across the evaluated WikiTQ-ST, TempTabQA-ST and SSTQA benchmarks. Table~\ref{tab:experiment} shows \oursys achieves the highest accuracy, exceeding the second-best method by 10.23\% on SSTQA benchmark.
This consistent outperformance can be attributed to three folds.

First, \oursys leverages the HO-Tree to represent \semitables, enabling explicit structural modeling while decoupling layout understanding from question answering. This design allows the general-purpose \llm to operate without directly parsing complex table layouts. 
Instead, given JSON-formatted header information, \oursys generates atomic operations and executes them on the HO-Tree for relevant data retrieval. In contrast, most methods, excluding vision-language models (VLMs) which can directly perceive layout information from table images, struggle to capture two-dimensional semantics using linear text representations.

Second, \oursys incorporates a novel question decomposition mechanism that breaks complex queries into simpler sub-questions, followed by precise operation-table alignment. This improves both operation generation accuracy and execution reliability, thereby improving overall question answering performance.

Third, \oursys dynamically combines top-down and bottom-up retrieval strategies based on question characteristics, enabling robust handling of diverse \semiqa scenarios. When the top-down retrieval fails or a question lacks explicit header references, bottom-up retrieval is employed. This flexible approach allows the system to effectively navigate complex table structures, outperforming other methods in challenging scenarios.

\revisionb{Additionally, we observe performance increase (around 3\%) even on the simple tables of WikiTQ-ST and TempTabQA-ST. For WikiTQ, where the majority of tables are fully structured, \oursys outperforms the second-best model by around 2\%. This modest improvement reflects \oursys's specialization for semi-structured tables with complex nested hierarchies. In contrast, on TempTabQA, where all tables are semi-structured but exhibit only shallow nesting and small sizes, \oursys achieves around 3\% improvement over the second-best approach. While our model could effectively models such structures, the overall retrieval pipeline remains relatively simple, limiting the performance gap over LLMs like Deepseek-V3 which can directly interpreting relevant information.}

Meanwhile, the experimental results highlight significant performance variations among the methods. NL2SQL-based approaches perform the worst on the SSTQA and TempTabQA dataset but achieve better results on WikiTQ, as the SQL generation paradigm is ill-suited for non-relational data, making it ineffective for \semitables.
TableLLM ranks second-lowest on SSTQA due to two main limitations: (1) its training is restricted to structured datasets, reducing its generalizability to semi-structured formats, and (2) it struggles with large-scale tables and complex layouts due to limited context length and one-dimensional semantic reasoning. Its improved performance on WikiTQ can be attributed to task-specific fine-tuning on this dataset.
Agent-based methods perform better on SSTQA due to their integration of external tools, but they fail to operate directly on semi-structured tables. The transformation of semi-structured data into structured formats results in the loss of critical layout information, reducing their effectiveness. 
TAT-LLM exhibits unexpectedly strong performance on the TempTabQA dataset. We attribute this to its fine-tuning on a large volume of financial data, which shares similarities with the temporal question types prevalent in TempTabQA, thereby contributing to its effectiveness.
Vision-language models excel at layout recognition through visual encoding but underperform in text-dense scenarios due to limited textual comprehension, especially in tables requiring strong semantic understanding.
Foundation models, while not explicitly designed for table-related tasks, achieve the second-best performance on SSTQA dataset. This is attributed to their robust contextual reasoning and semantic interpretation abilities, enabling accurate answer inference especially when tabular layout understanding is secondary to interpreting textual content.

\hi{Accuracy under Different Table Difficulties.}  We categorize tables in SSTQA benchmark into three levels of difficulty (i.e., Simple, Medium, Hard) by layout complexity and content length. Figure~\ref{fig:table_difficulty} presents the comparative performance evaluation across these difficulty levels. Three key observations emerge from our analysis. 

First, both \oursys and foundation models exhibit progressively decreasing performance as table difficulty increases, underscoring the inherent challenges posed by complex layouts and large-scale \semitables. In contrast, other methods demonstrate only marginal performance variation across difficulty levels. 
We posit that these models primarily address questions with less table structure comprehension.
Consequently, performance differences among them are largely driven by architectural variations rather than structural modeling capabilities.

Second, although \oursys shows a modest performance decline on hard-level tables, it consistently outperforms all methods by a substantial margin (e.g., exceeding the second-best model by over 20\% on the SSTQA dataset). The reasons are three-fold: (1) the hierarchical HO-Tree representation, which facilitates efficient processing of large tables; (2) the question decomposition mechanism, which simplifies complex queries into tractable sub-questions; and (3) the operation-table alignment strategy, which ensures accurate and context-aware data retrieval.

Third, performance differences across models are less pronounced. This can be attributed to three key factors: (1) the smaller table sizes, which reduce structural complexity; (2) the predominance of semantically driven questions that require less explicit layout reasoning; and (3) the dataset’s focus on temporal question answering within a single scenario, which limits question diversity and diminishes the impact of advanced structural modeling.

\hi{Analysis on Table + QA Difficulty.} \revisiona{We categorize the Table+QA tasks into three difficulty levels. As shown in Table~\ref{tab:qa_diff}, \oursys outperforms all baselines across these levels. While both foundation models and \oursys perform well on simple cases, accuracy drops as difficulty increases. Notably, \oursys demonstrates superior performance on hard cases, attributed to its HO-Tree-based representation and operation-pipeline-driven QA strategy.}

We categorize Table+QA tasks into three difficulty levels (Table~\ref{tab:qa_diff}). \oursys consistently outperforms all baselines across these levels. While both foundation models and \oursys perform well on simple cases, accuracy drops as difficulty increases. Notably, \oursys excels on hard cases, benefiting from its HO-Tree representation and operation-pipeline-driven QA strategy.


\begin{table}[!t]
    \small
    \centering
    \setlength{\tabcolsep}{6pt}
    \caption{\revisiona{Analysis of Table + QA Difficulty on SSTQA.}}
    \vspace{-0.4cm}
    \begin{tabular}{ l c c c }
        \hline
        Methods (Acc) & Simple & Medium & Hard \\
        \hline
        DeepseekV3 & 92.94\% & 61.83\% & 47.19\% \\
        GPT-4o & 86.64\% & 59.75\% & 43.26\% \\
        \oursys & \textbf{93.97\%} & \textbf{62.66\%} & \textbf{58.43\%} \\
        \hline
    \end{tabular}
    \vspace{-.5cm}
    \label{tab:qa_diff}
\end{table}

\vspace{-0.5em}
\subsection{Fine-grained Analysis} \label{sec:sec:fine_grained_analysis}

In this section, we discuss the quality of meta-information detection, analyze question answering latency, and examine the impact of pipeline errors.

\hi{Quality of Meta Information Detection.}
\revisiona{We evaluate the Table2Tree module's accuracy in converting \semitable into HO-Tree. Experiment results show that the untuned VLM achieves 93.14\% on SSTQA, 94.32\% on TempTabQA and 92.31\% on WikiTQ, which is sufficiently high for accurate HO-Tree construction.}


\hi{Analysis of Backward Verification.}

\revisionc{To assess the potential negative impact of generating suboptimal question alternatives on question answering accuracy, we quantify the number of such bad alternatives and their corresponding answers on the SSTQA dataset. Experimental results show a false negative rate of 4.78\% under the few-shot learning setting, indicating that misjudgments in backward verification have minimal impact on table QA performance.}

\hi{Latency Analysis.}
\revisionc{The runtime of \oursys is primarily influenced by the cost of accessing the LLM, largely due to network latency. As \oursys performs question answering via pipeline-based operation generation, the runtime per query is inherently unstable. \oursys requires around 30 seconds per question (ignoring bias caused by factors like network communications), with 2.89 pipeline operations on average. This is substantially faster than the agent-based method, which incurs higher latency due to a greater number of operations with more API calls, and slightly slower than the fine-tuning approach, which benefits from local deployment and direct reasoning.}

\hi{Effects of Pipeline Mistakes.}
\hi{Pipeline Mistakes Analysis.} \revisiona{Mistakes in the \oursys pipeline primarily arise from two sources. First, mistakes in meta-information detection by the VLM can lead to incorrect HO-Tree representations, resulting in more complex data retrieval paths (e.g., locating related data dispersed across different subtrees) and reduced overall efficiency (e.g., from 10 to 40 seconds). Second, semantic misinterpretations by the LLM, such incorrectly splitting a combined address-phone entry into separate fields, can trigger unnecessary lookups and potentially yield incorrect answers, leading to additional verification and iteration, and thereby diminishing efficiency.}

\begin{table}[!t]
    \small
    \centering
    \setlength{\tabcolsep}{6pt}
    \caption{Ablation Study on \oursys Modules}
    \vspace{-0.3cm}
    \begin{tabular}{ l c c }
        \hline
        \multirow{2}{*}{Model} & \multicolumn{2}{c}{SSTQA} \\
        & Acc & ROUGE-L \\
        \hline
        Full Model (DeepseekV3) & 72.39\% & 52.19\% \\
        \hline
        GPT-4o & 62.12\% & 43.86\% \\
        DeepseekV3 & 62.26\% & 46.17\% \\
        \hline
        w/o Table2Tree & 57.24\%(-15.15\%) & 41.55\%(-10.64\%) \\
        w/o Question Decomposition & 68.06\%(-4.33\%) & 48.09\%(-4.10\%) \\
        w/o Operation-Table Alignment & 71.07\%(-1.32\%) & 50.86\%(-1.33\%) \\
        w/o Data Manipulation Operation & 65.09\%(-7.30\%) & 47.13\%(-5.06\%) \\
        w/o Answer Verifier & 66.10\%(-6.29\%) & 47.46\%(-4.73\%) \\
        \hline
    \end{tabular}
    \vspace{-.7cm}
    \label{tab:ablation}
\end{table}

\begin{table*}[!t]
    \footnotesize
    \centering
    \vspace{-1.75em}
    \setlength{\tabcolsep}{2pt}
    \caption{Case Study on SSTQA Dataset.}
    \vspace{-0.1cm}
    \begin{tabular}{ c c l l c c c c c }
        \hline
        \makecell{Table\\Id} & \makecell{Table\\Difficulty} & Layout Representation & Question  & TableLLaMA & ReAcTable & \makecell{mPLUG-\\DocOwl1.5} & GPT-4o & \oursys \\
        \hline
        5 & Simple & $L.4 \rightarrow L.3 \rightarrow[L.2_1, \dots, L.2_{12}]$ & \makecell[l]{Summarize the reimbursement\\ activities of Tian Xiaohong.} & \XSolidBrush & \XSolidBrush & \XSolidBrush & \XSolidBrush & \XSolidBrush \\
        \hline
        19 & Simple & $L.4 \rightarrow L.3 \rightarrow [L.2_1, \dots, L.2_{3}]$ & \makecell[l]{What are the components of employee compensation?} & \XSolidBrush & \XSolidBrush & \Checkmark & \Checkmark & \Checkmark \\
        \hline
        15 & Simple & $L.4 \rightarrow L.3 \rightarrow[L.2_1, L.2_2, L.2_3]$ & \makecell[l]{What documents must a Continuity and \\Availability Planner submit to the IT\\ Service Management Committee?} & \XSolidBrush & \Checkmark & \Checkmark & \Checkmark & \Checkmark \\
        \hline
        20 & Simple & \makecell[l]{$L.4 \rightarrow L.3 \rightarrow [\{L.4 $\\$ \rightarrow L.3 \rightarrow [L.2_1, \dots, L.2_{6}]\}_1, $\\$ \{L.4 \rightarrow [L.2_1, \dots, L.2_{6}]\}_2]$} & \makecell[l]{What are the categories of variable\\ manufacturing overhead?} & \Checkmark & \XSolidBrush & \XSolidBrush & \Checkmark & \Checkmark \\
        \hline
        4 & Medium & \makecell[l]{$L.3 \rightarrow [\{L.4 \rightarrow [L.2_1, L.2_2, L.2_3]\}_1, $\\$ \dots, \{L.4 \rightarrow [L.2_1, L.2_2, L.2_3]_6\}]$} & \makecell[l]{How many items are there in \\drawing technology?} & \XSolidBrush & \XSolidBrush & \XSolidBrush & \XSolidBrush & \Checkmark \\
        \hline
        95 & Medium & $L.4 \rightarrow L.3 \rightarrow[L.2_1, \dots, L.2_8]$ & \makecell[l]{Which employees in the table have 18\\ years of service?} & \XSolidBrush & \XSolidBrush & \XSolidBrush & \XSolidBrush & \Checkmark \\
        \hline
        100 & Medium & \makecell[l]{$L.4 \rightarrow \{L.4 \rightarrow L.3 \rightarrow $\\$[L.1_1, L.1_2, \dots, L.1_{34}]\}$} & \makecell[l]{What is the net cash flow generated\\ from investing activities?} & \Checkmark & \Checkmark & \Checkmark & \XSolidBrush & \Checkmark \\
        \hline
        87 & Medium & $L.4 \rightarrow L.3 \rightarrow[L.2_1, \dots, L.2_{10}]$ & \makecell[l]{What are the evaluation criteria for\\ work attitude?} & \XSolidBrush & \XSolidBrush & \XSolidBrush & \Checkmark & \Checkmark \\
        \hline
        1 & Hard & \makecell[l]{$L.4 \rightarrow [L.1, \{L.4 \rightarrow \{L.3 $\\$ \rightarrow [L.1_1, L.1_2]\}\}_1, \{L.4 \rightarrow \{L.3 $\\$ \rightarrow [L.2_1, \dots, L.2_4]\}\}_2, \dots]$} & \makecell[l]{How many secondary indicators are\\ included under the performance\\ metric's efficiency indicators?} & \XSolidBrush & \XSolidBrush & \XSolidBrush & \XSolidBrush & \Checkmark \\
        \hline
        10 & Hard & \makecell[l]{$L.3 \rightarrow [\{L.4 \rightarrow [L.2_1, L.2_2]\}_1, $\\$ \dots, \{L.4 \rightarrow [L.2_1, L.2_2]_4\}]$} & \makecell[l]{How many phases are included in the\\ Change Phase Code Table?} & \XSolidBrush & \XSolidBrush & \XSolidBrush & \Checkmark & \Checkmark \\
        \hline
        91 & Hard & $L.4 \rightarrow L.3 \rightarrow[L.2_1, \dots, L.2_{8}]$ & \makecell[l]{What are the beginning and ending\\ balances of total assets?} & \XSolidBrush & \XSolidBrush & \XSolidBrush & \Checkmark & \Checkmark \\
        \hline
        30 & Hard & $L.3 \rightarrow[L.1, L.2_1, \dots, L.2_{6}]$ & \makecell[l]{How many categories are there for service\\ sub items with service number 'XX-R-I-4'?} & \XSolidBrush & \XSolidBrush & \XSolidBrush & \XSolidBrush & \Checkmark \\
        \hline
    \end{tabular}
    \label{tab:case_study}
    \vspace{-0.3cm}
\end{table*}

\vspace{-0.5em}
\subsection{Ablation Study on \oursys Modules} \label{sec:sec:ablation_study}

In this section, we perform an ablation study on \oursys from five perspectives. Results are reported in Table~\ref{tab:ablation}.

\hi{Without Table2Tree.}
\revisionc{We disable the Table2Tree module and instead apply \oursys directly to raw \semitables, which evaluates the importance of explicit table layout modeling. The removal leads to the most significant degradation (an absolute accuracy drop of 15.15\%), demonstrating the critical role of HO-Tree-based structural representation in handling complex \semitables. This also highlights that foundation models alone struggle to capture intricate layout semantics without explicit structural guidance.}

\hi{Without Question Decomposition.}
We remove the question decomposition module, requiring \oursys to process complex queries in a single step. This results in a 4.33\% accuracy drop, confirming the necessity of decomposition for effective multi-hop reasoning. Without decomposition, the \oursys fails to isolate intermediate steps, leading to compounding errors in reasoning chains.

\hi{Without Operation-Table Alignment.}
We omit the operation-table alignment mechanism to test whether the LLM in \oursys can inherently align operations with table content. A 1.32\% performance decline is observed, indicating that while LLMs possess semantic reasoning ability, explicit alignment could still improve execution precision. This suggests that structural grounding remains beneficial even for advanced models with strong language understanding capabilities.

\hi{Without Data Manipulation Operations.}
We restrict \oursys to data retrieval, alignment, and reasoning operations, disabling data manipulation functions. This leads to a 7.30\% accuracy drop, underscoring the frequent necessity of manipulation operations and validating the completeness of our atomic operation set. Many questions inherently require operations such as filtering and calculation, which cannot be bypassed through reasoning alone.

\hi{Without Answer Verifier.}
To evaluate the impact of self-verification, we remove the answer verifier module. Accuracy drops by 6.29\%, suggesting that the verifier plays a vital role in detecting and correcting execution errors, thereby enhancing output reliability. This module is especially useful when wrong intermediate results or final answer are generated during multi-step execution.


\begin{sloppypar}
Collectively, these results demonstrate that each module in \oursys addresses distinct yet complementary challenges in \semiqa, and their synergistic integration is vital for effectively tackling the layout-intensive questions in SSTQA.
\end{sloppypar}



\vspace{-0.5em}
\subsection{Case Study on SSTQA Dataset} \label{sec:sec:case_study}

For each table difficulty level in the SSTQA dataset, we select four representative question-answering cases. Table~\ref{tab:case_study} presents the abstract \semitable layout representations, the corresponding questions, and the results from five selected methods. The layout representations follow the table definitions introduced in Section~\ref{sec:overview}, where $L.1$ denotes a Header-Single-Value structure, $L.2$ denotes Header-Multiple-Value, $L.3$ denotes Orthogonal Tables, and $L.4$ denotes Header-Orthogonal-Tables. A rightward arrow indicates the construction of a layout. For example, $L.4 \rightarrow L.2$ indicates that headers are added to the $L.2$ layout. 

Two key observations emerge from the Table~\ref{tab:case_study}: (1) In terms of structural complexity, tables with complex layouts (e.g., Tables 20, 4, 1, 10) often lead to errors for most methods except ST-Raptor. (2) Regarding questions, those requiring math operations (e.g., Tables 4, 1, 30) demand a deeper understanding of table structure, where \oursys consistently excels other methods.

\revisionc{Regarding limitations, \oursys may occasionally irregular layout patterns, such as erroneously treating horizontally merged content cells as headers, which can negatively affect the QA accuracy. }
\revisionc{Besides, both \oursys and the baselines face challenges in decomposing questions involving complex pipelines (e.g., resembling multi-level nested {SQL} queries), requiring techniques such as specialized \llm fine-tuning. Nonetheless, such cases are infrequent in semi-structured table QA tasks.}



\vspace{-1em}
\section{Related Works} \label{sec:related_works}

\subsection{Structured Table QA.} Mainstream approaches can be categorized into NL2SQL, NL2Code and vision-language model based methods. NL2SQL~\cite{PETSQL,XiYanSQL,OpenSearchSQL} focuses on translating natural language queries into structured SQL commands by leveraging techniques such as (1) schema linking, which aligns user intents with database schema to resolve ambiguities, and (2) content retrieval, which dynamically extracts relevant information from the database to refine query generation. VLM based methods~\cite{TableLLaVA, DocOwl1.5} transform the table into image for analysis and question answering.

\vspace{-1.em}
\subsection{Semi-Structured Table QA.} Semi-structured tables bring a large challenge for table understanding and render traditional Text2SQL strategy ineffective. To address the issue, numerous excellent research efforts have been carried out ~\cite{NeuralQuestion,SemiKnowledge,GrabTable,Compositional}. For instance, Wang et al. proposed an end-to-end system~\cite{NeuralQuestion} that uses semi-structured tables as knowledge sources by first finding the most similar tables and then selecting the most relevant table cells to derive the answer.  Additionally, Liu et al. proposed the GrabTab method~\cite{GrabTable} featuring a Component Deliberator that efficiently leverages multiple table components without requiring complex post-processing, for addressing the challenge of recognizing complex and irregular table structures. \revisionb{Inspired by NL2SQL techniques, Lu et al.~\cite{lu2025bridge} propose to convert natural language queries into NoSQL ones, but only produce intermediate results, lacking the end-to-end semi-structured table QA capability like \oursys.} Moreover, Gupta et al. built a TEMPTABQA dataset~\cite{TEMPTAQA} from 1,208 Wikipedia Infobox tables to evaluate the temporal reasoning capabilities, and found that even top-performing LLMs fall behind human performance by over 13.5 F1 points. \revisionb{Some works conduct information extraction over semi-structured data. TWIX~\cite{TWIX} assumes that many semi-structured data is generate from one similar layout template and proposes a method that first reconstruct the template than extract the content. However, it lacks support for merged cells and cannot be readily transformed into HO-Trees within our problem scope.} \revisionb{Another approach is convert the semi-structured data into structured formats for downstream analysis~\cite{arora2025semi2stru}. However, this conversion process can introduce information loss and reduce answer accuracy.} Overall, these findings highlight that despite significant advancements, substantial challenges remain in effectively addressing semi-structured table QA.

\vspace{-0.3em}
\subsection{In-Context Table QA.} Although table-based reasoning has shown remarkable progress with the emergence of LLMs~\cite{deng-etal-2024-tables}, table QA solutions encounter significant performance degradation when confronted with the rich and diverse evidence present in tables. To enable LLMs to sufficiently understand both the tables and question information, decomposition plays a pivotal role in table QA~\cite{ToolAssisted,LLMDecomposers,ReAcTable,PETSQL,DecomposedPrompting}. Ye et al.~\cite{LLMDecomposers} introduces a method that effectively leverages LLMs to decompose large tables into relevant sub-tables and complex questions into simpler sub-questions through in-context prompting. In addition to in-context learning from data sources, the ReAcTable framework~\cite{ReAcTable} proposed by Zhang et al. aims to improve complex table QA performance by incorporating execution feedback. This feedback mechanism, rooted in the ReAct framework, enables the system to dynamically adjust its operations based on the results of previous actions and addresses challenges such as interpreting complex data semantics, handling generated errors, and performing intricate data transformations.

\vspace{+0.6em}
\section{Conclusion} \label{sec:conclusion}


In this paper, we introduced \oursys, a tree-based framework aimed at addressing the critical challenges of automating question answering over semi-structured tables. Central to our approach is the Hierarchical Orthogonal Tree (HO-Tree), a formal representation capable of capturing complex table layouts, including hierarchical headers, merged cells, and implicit relationships. We designed a set of basic tree operations over HO-Trees to enable LLMs to perform layout-aware tasks. Given a user question, \oursys decomposes it into simpler subquestions, constructs corresponding tree-operation pipelines, and executes them to retrieve relevant information or derive the final answer. To ensure both execution correctness and answer reliability, we proposed a two-stage verification mechanism combining forward constraint checking and backward answer validation. Additionally, we constructed the SSTQA benchmark, consisting of 764 questions over 102 real-world semi-structured tables. Experimental results demonstrate that \oursys outperforms all baselines by up to 20\% in answer accuracy.


\clearpage
\newpage
\bibliographystyle{ACM-Reference-Format}
\bibliography{ref}


\begin{thebibliography}{48}


\ifx \showCODEN    \undefined \def \showCODEN     #1{\unskip}     \fi
\ifx \showDOI      \undefined \def \showDOI       #1{#1}\fi
\ifx \showISBNx    \undefined \def \showISBNx     #1{\unskip}     \fi
\ifx \showISBNxiii \undefined \def \showISBNxiii  #1{\unskip}     \fi
\ifx \showISSN     \undefined \def \showISSN      #1{\unskip}     \fi
\ifx \showLCCN     \undefined \def \showLCCN      #1{\unskip}     \fi
\ifx \shownote     \undefined \def \shownote      #1{#1}          \fi
\ifx \showarticletitle \undefined \def \showarticletitle #1{#1}   \fi
\ifx \showURL      \undefined \def \showURL       {\relax}        \fi
\providecommand\bibfield[2]{#2}
\providecommand\bibinfo[2]{#2}
\providecommand\natexlab[1]{#1}
\providecommand\showeprint[2][]{arXiv:#2}

\bibitem[\protect\citeauthoryear{??}{Fin}{[n.d.]}]%
        {FinanceData}
 \bibinfo{year}{[n.d.]}\natexlab{}.
\newblock
\newblock
\urldef\tempurl%
\url{https://www.frontiersin.org/research-topics/21489/knowledge-discovery-from-unstructured-data-in-finance}
\showURL{%
\tempurl}


\bibitem[\protect\citeauthoryear{??}{Med}{[n.d.]}]%
        {MedicalData}
 \bibinfo{year}{[n.d.]}\natexlab{}.
\newblock
\newblock
\urldef\tempurl%
\url{https://enterprises.upmc.com/resources/insights/health-cares-unstructured-data-challenge/}
\showURL{%
\tempurl}


\bibitem[\protect\citeauthoryear{??}{Com}{[n.d.]}]%
        {CommerceData}
 \bibinfo{year}{[n.d.]}\natexlab{}.
\newblock
\newblock
\urldef\tempurl%
\url{https://pages.cs.wisc.edu/~jbeckham/TR/cnet.pdf}
\showURL{%
\tempurl}


\bibitem[\protect\citeauthoryear{Abiteboul}{Abiteboul}{1997}]%
        {SemiData}
\bibfield{author}{\bibinfo{person}{Serge Abiteboul}.} \bibinfo{year}{1997}\natexlab{}.
\newblock \showarticletitle{Querying semi-structured data}. In \bibinfo{booktitle}{\emph{Database Theory --- ICDT '97}}, \bibfield{editor}{\bibinfo{person}{Foto Afrati} {and} \bibinfo{person}{Phokion Kolaitis}} (Eds.). \bibinfo{publisher}{Springer Berlin Heidelberg}, \bibinfo{address}{Berlin, Heidelberg}, \bibinfo{pages}{1--18}.
\newblock
\showISBNx{978-3-540-49682-3}


\bibitem[\protect\citeauthoryear{Alwateer, Atlam, Abd El-Raouf, Ghoneim, and Gad}{Alwateer et~al\mbox{.}}{2024}]%
        {alwateer2024missing}
\bibfield{author}{\bibinfo{person}{Majed Alwateer}, \bibinfo{person}{El-Sayed Atlam}, \bibinfo{person}{Mahmoud~Mohammed Abd El-Raouf}, \bibinfo{person}{Osama~A. Ghoneim}, {and} \bibinfo{person}{Ibrahim Gad}.} \bibinfo{year}{2024}\natexlab{}.
\newblock \showarticletitle{Missing Data Imputation: A Comprehensive Review}.
\newblock \bibinfo{journal}{\emph{Journal of Computer and Communications}} \bibinfo{volume}{12}, \bibinfo{number}{11} (\bibinfo{year}{2024}), \bibinfo{pages}{53--75}.
\newblock
\urldef\tempurl%
\url{https://doi.org/10.4236/jcc.2024.1211004}
\showDOI{\tempurl}


\bibitem[\protect\citeauthoryear{Arora, Yang, Eyuboglu, Narayan, Hojel, Trummer, and Ré}{Arora et~al\mbox{.}}{2025}]%
        {arora2025semi2stru}
\bibfield{author}{\bibinfo{person}{Simran Arora}, \bibinfo{person}{Brandon Yang}, \bibinfo{person}{Sabri Eyuboglu}, \bibinfo{person}{Avanika Narayan}, \bibinfo{person}{Andrew Hojel}, \bibinfo{person}{Immanuel Trummer}, {and} \bibinfo{person}{Christopher Ré}.} \bibinfo{year}{2025}\natexlab{}.
\newblock \bibinfo{title}{Language Models Enable Simple Systems for Generating Structured Views of Heterogeneous Data Lakes}.
\newblock
\newblock
\showeprint[arxiv]{2304.09433}~[cs.CL]
\urldef\tempurl%
\url{https://arxiv.org/abs/2304.09433}
\showURL{%
\tempurl}


\bibitem[\protect\citeauthoryear{Barboule, Piwowarski, and Chabot}{Barboule et~al\mbox{.}}{2025}]%
        {barboule2025survey}
\bibfield{author}{\bibinfo{person}{Camille Barboule}, \bibinfo{person}{Benjamin Piwowarski}, {and} \bibinfo{person}{Yoan Chabot}.} \bibinfo{year}{2025}\natexlab{}.
\newblock \bibinfo{title}{Survey on Question Answering over Visually Rich Documents: Methods, Challenges, and Trends}.
\newblock
\newblock
\showeprint[arxiv]{2501.02235}~[cs.CL]
\urldef\tempurl%
\url{https://arxiv.org/abs/2501.02235}
\showURL{%
\tempurl}


\bibitem[\protect\citeauthoryear{Borchmann and Wydmuch}{Borchmann and Wydmuch}{2025}]%
        {BorchmannQnC}
\bibfield{author}{\bibinfo{person}{\L{}ukasz Borchmann} {and} \bibinfo{person}{Marek Wydmuch}.} \bibinfo{year}{2025}\natexlab{}.
\newblock \bibinfo{title}{Query and Conquer: Execution-Guided SQL Generation}.
\newblock
\newblock
\showeprint[arxiv]{2503.24364}~[cs.CL]
\urldef\tempurl%
\url{https://arxiv.org/abs/2503.24364}
\showURL{%
\tempurl}


\bibitem[\protect\citeauthoryear{Burdick, Danilevsky, Evfimievski, Katsis, and Wang}{Burdick et~al\mbox{.}}{2020}]%
        {Burdick2020Table}
\bibfield{author}{\bibinfo{person}{Douglas Burdick}, \bibinfo{person}{Marina Danilevsky}, \bibinfo{person}{Alexandre~V. Evfimievski}, \bibinfo{person}{Yannis Katsis}, {and} \bibinfo{person}{Nancy Wang}.} \bibinfo{year}{2020}\natexlab{}.
\newblock \showarticletitle{Table Extraction and Understanding for Scientific and Enterprise Applications}.
\newblock \bibinfo{journal}{\emph{Proc. VLDB Endow.}} \bibinfo{volume}{13}, \bibinfo{number}{12} (\bibinfo{year}{2020}), \bibinfo{pages}{3433--3436}.
\newblock
\urldef\tempurl%
\url{https://doi.org/10.14778/3415478.3415563}
\showDOI{\tempurl}


\bibitem[\protect\citeauthoryear{Chen, Chen, Koudas, and Yu}{Chen et~al\mbox{.}}{2025a}]%
        {ChenRTS}
\bibfield{author}{\bibinfo{person}{Kaiwen Chen}, \bibinfo{person}{Yueting Chen}, \bibinfo{person}{Nick Koudas}, {and} \bibinfo{person}{Xiaohui Yu}.} \bibinfo{year}{2025}\natexlab{a}.
\newblock \showarticletitle{Reliable Text-to-SQL with Adaptive Abstention}.
\newblock \bibinfo{journal}{\emph{Proceedings of the ACM on Management of Data}} \bibinfo{volume}{3}, \bibinfo{number}{1} (\bibinfo{date}{Feb.} \bibinfo{year}{2025}), \bibinfo{pages}{69:1--69:30}.
\newblock
\urldef\tempurl%
\url{https://doi.org/10.1145/3709719}
\showDOI{\tempurl}


\bibitem[\protect\citeauthoryear{Chen, Wang, Cao, Liu, Gao, Cui, Zhu, Ye, Tian, Liu, Gu, Wang, Li, Ren, Chen, Luo, Wang, Jiang, Wang, He, Shi, Zhang, Lv, Wang, Shao, Chu, Tu, He, Wu, Deng, Ge, Chen, Zhang, Wang, Dou, Lu, Zhu, Lu, Lin, Qiao, Dai, and Wang}{Chen et~al\mbox{.}}{2025b}]%
        {InterVL}
\bibfield{author}{\bibinfo{person}{Zhe Chen}, \bibinfo{person}{Weiyun Wang}, \bibinfo{person}{Yue Cao}, \bibinfo{person}{Yangzhou Liu}, \bibinfo{person}{Zhangwei Gao}, \bibinfo{person}{Erfei Cui}, \bibinfo{person}{Jinguo Zhu}, \bibinfo{person}{Shenglong Ye}, \bibinfo{person}{Hao Tian}, \bibinfo{person}{Zhaoyang Liu}, \bibinfo{person}{Lixin Gu}, \bibinfo{person}{Xuehui Wang}, \bibinfo{person}{Qingyun Li}, \bibinfo{person}{Yimin Ren}, \bibinfo{person}{Zixuan Chen}, \bibinfo{person}{Jiapeng Luo}, \bibinfo{person}{Jiahao Wang}, \bibinfo{person}{Tan Jiang}, \bibinfo{person}{Bo Wang}, \bibinfo{person}{Conghui He}, \bibinfo{person}{Botian Shi}, \bibinfo{person}{Xingcheng Zhang}, \bibinfo{person}{Han Lv}, \bibinfo{person}{Yi Wang}, \bibinfo{person}{Wenqi Shao}, \bibinfo{person}{Pei Chu}, \bibinfo{person}{Zhongying Tu}, \bibinfo{person}{Tong He}, \bibinfo{person}{Zhiyong Wu}, \bibinfo{person}{Huipeng Deng}, \bibinfo{person}{Jiaye Ge}, \bibinfo{person}{Kai Chen}, \bibinfo{person}{Kaipeng Zhang}, \bibinfo{person}{Limin
  Wang}, \bibinfo{person}{Min Dou}, \bibinfo{person}{Lewei Lu}, \bibinfo{person}{Xizhou Zhu}, \bibinfo{person}{Tong Lu}, \bibinfo{person}{Dahua Lin}, \bibinfo{person}{Yu Qiao}, \bibinfo{person}{Jifeng Dai}, {and} \bibinfo{person}{Wenhai Wang}.} \bibinfo{year}{2025}\natexlab{b}.
\newblock \bibinfo{title}{Expanding Performance Boundaries of Open-Source Multimodal Models with Model, Data, and Test-Time Scaling}.
\newblock
\newblock
\showeprint[arxiv]{2412.05271}~[cs.CV]
\urldef\tempurl%
\url{https://arxiv.org/abs/2412.05271}
\showURL{%
\tempurl}


\bibitem[\protect\citeauthoryear{Codd}{Codd}{1970}]%
        {StructuredData}
\bibfield{author}{\bibinfo{person}{E.~F. Codd}.} \bibinfo{year}{1970}\natexlab{}.
\newblock \showarticletitle{A relational model of data for large shared data banks}.
\newblock \bibinfo{journal}{\emph{Commun. ACM}} \bibinfo{volume}{13}, \bibinfo{number}{6} (\bibinfo{date}{June} \bibinfo{year}{1970}), \bibinfo{pages}{377–387}.
\newblock
\showISSN{0001-0782}
\urldef\tempurl%
\url{https://doi.org/10.1145/362384.362685}
\showDOI{\tempurl}


\bibitem[\protect\citeauthoryear{DeepSeek-AI}{DeepSeek-AI}{2025}]%
        {deepseekr1}
\bibfield{author}{\bibinfo{person}{DeepSeek-AI}.} \bibinfo{year}{2025}\natexlab{}.
\newblock \bibinfo{title}{DeepSeek-R1: Incentivizing Reasoning Capability in LLMs via Reinforcement Learning}.
\newblock
\newblock
\showeprint[arxiv]{2501.12948}~[cs.CL]
\urldef\tempurl%
\url{https://arxiv.org/abs/2501.12948}
\showURL{%
\tempurl}


\bibitem[\protect\citeauthoryear{DeepSeek-AI, Liu, Feng, et~al\mbox{.}}{DeepSeek-AI et~al\mbox{.}}{2025}]%
        {DeepSeekV3}
\bibfield{author}{\bibinfo{person}{DeepSeek-AI}, \bibinfo{person}{Aixin Liu}, \bibinfo{person}{Bei Feng}, {et~al\mbox{.}}} \bibinfo{year}{2025}\natexlab{}.
\newblock \bibinfo{title}{DeepSeek-V3 Technical Report}.
\newblock
\newblock
\showeprint[arxiv]{2412.19437}~[cs.CL]
\urldef\tempurl%
\url{https://arxiv.org/abs/2412.19437}
\showURL{%
\tempurl}


\bibitem[\protect\citeauthoryear{Deng, Sun, He, Sikka, Chen, Ma, Zhang, and Mihalcea}{Deng et~al\mbox{.}}{2024}]%
        {deng-etal-2024-tables}
\bibfield{author}{\bibinfo{person}{Naihao Deng}, \bibinfo{person}{Zhenjie Sun}, \bibinfo{person}{Ruiqi He}, \bibinfo{person}{Aman Sikka}, \bibinfo{person}{Yulong Chen}, \bibinfo{person}{Lin Ma}, \bibinfo{person}{Yue Zhang}, {and} \bibinfo{person}{Rada Mihalcea}.} \bibinfo{year}{2024}\natexlab{}.
\newblock \showarticletitle{Tables as Texts or Images: Evaluating the Table Reasoning Ability of {LLM}s and {MLLM}s}. In \bibinfo{booktitle}{\emph{Findings of the Association for Computational Linguistics: ACL 2024}}. \bibinfo{publisher}{Association for Computational Linguistics}, \bibinfo{address}{Bangkok, Thailand}, \bibinfo{pages}{407--426}.
\newblock
\urldef\tempurl%
\url{https://doi.org/10.18653/v1/2024.findings-acl.23}
\showDOI{\tempurl}


\bibitem[\protect\citeauthoryear{Gao, Liu, Li, Shi, Zhu, Wang, Li, Li, Hong, Luo, Gao, Mou, and Li}{Gao et~al\mbox{.}}{2025}]%
        {XiYanSQL}
\bibfield{author}{\bibinfo{person}{Yingqi Gao}, \bibinfo{person}{Yifu Liu}, \bibinfo{person}{Xiaoxia Li}, \bibinfo{person}{Xiaorong Shi}, \bibinfo{person}{Yin Zhu}, \bibinfo{person}{Yiming Wang}, \bibinfo{person}{Shiqi Li}, \bibinfo{person}{Wei Li}, \bibinfo{person}{Yuntao Hong}, \bibinfo{person}{Zhiling Luo}, \bibinfo{person}{Jinyang Gao}, \bibinfo{person}{Liyu Mou}, {and} \bibinfo{person}{Yu Li}.} \bibinfo{year}{2025}\natexlab{}.
\newblock \bibinfo{title}{A Preview of XiYan-SQL: A Multi-Generator Ensemble Framework for Text-to-SQL}.
\newblock
\newblock
\showeprint[arxiv]{2411.08599}~[cs.AI]
\urldef\tempurl%
\url{https://arxiv.org/abs/2411.08599}
\showURL{%
\tempurl}


\bibitem[\protect\citeauthoryear{Gupta, Kandoi, Vora, Zhang, He, Reinanda, and Srikumar}{Gupta et~al\mbox{.}}{2023a}]%
        {temptabqa}
\bibfield{author}{\bibinfo{person}{Vivek Gupta}, \bibinfo{person}{Pranshu Kandoi}, \bibinfo{person}{Mahek Vora}, \bibinfo{person}{Shuo Zhang}, \bibinfo{person}{Yujie He}, \bibinfo{person}{Ridho Reinanda}, {and} \bibinfo{person}{Vivek Srikumar}.} \bibinfo{year}{2023}\natexlab{a}.
\newblock \showarticletitle{{T}emp{T}ab{QA}: Temporal Question Answering for Semi-Structured Tables}. In \bibinfo{booktitle}{\emph{Proceedings of the 2023 Conference on Empirical Methods in Natural Language Processing}}, \bibfield{editor}{\bibinfo{person}{Houda Bouamor}, \bibinfo{person}{Juan Pino}, {and} \bibinfo{person}{Kalika Bali}} (Eds.). \bibinfo{publisher}{Association for Computational Linguistics}, \bibinfo{address}{Singapore}, \bibinfo{pages}{2431--2453}.
\newblock
\urldef\tempurl%
\url{https://doi.org/10.18653/v1/2023.emnlp-main.149}
\showDOI{\tempurl}


\bibitem[\protect\citeauthoryear{Gupta, Kandoi, Vora, Zhang, He, Reinanda, and Srikumar}{Gupta et~al\mbox{.}}{2023b}]%
        {TEMPTAQA}
\bibfield{author}{\bibinfo{person}{Vivek Gupta}, \bibinfo{person}{Pranshu Kandoi}, \bibinfo{person}{Mahek Vora}, \bibinfo{person}{Shuo Zhang}, \bibinfo{person}{Yujie He}, \bibinfo{person}{Ridho Reinanda}, {and} \bibinfo{person}{Vivek Srikumar}.} \bibinfo{year}{2023}\natexlab{b}.
\newblock \showarticletitle{{T}emp{T}ab{QA}: Temporal Question Answering for Semi-Structured Tables}. In \bibinfo{booktitle}{\emph{Proceedings of the 2023 Conference on Empirical Methods in Natural Language Processing}}, \bibfield{editor}{\bibinfo{person}{Houda Bouamor}, \bibinfo{person}{Juan Pino}, {and} \bibinfo{person}{Kalika Bali}} (Eds.). \bibinfo{publisher}{Association for Computational Linguistics}, \bibinfo{address}{Singapore}, \bibinfo{pages}{2431--2453}.
\newblock
\urldef\tempurl%
\url{https://doi.org/10.18653/v1/2023.emnlp-main.149}
\showDOI{\tempurl}


\bibitem[\protect\citeauthoryear{Herzig, Nowak, M{\"u}ller, Piccinno, and Eisenschlos}{Herzig et~al\mbox{.}}{2020}]%
        {herzig2020tapas}
\bibfield{author}{\bibinfo{person}{Jonathan Herzig}, \bibinfo{person}{Pawe{\l}~Krzysztof Nowak}, \bibinfo{person}{Thomas M{\"u}ller}, \bibinfo{person}{Francesco Piccinno}, {and} \bibinfo{person}{Julian~Martin Eisenschlos}.} \bibinfo{year}{2020}\natexlab{}.
\newblock \showarticletitle{TAPAS: Weakly Supervised Table Parsing via Pre-training}. In \bibinfo{booktitle}{\emph{Proceedings of the 58th Annual Meeting of the Association for Computational Linguistics (ACL)}}. \bibinfo{publisher}{Association for Computational Linguistics}, \bibinfo{pages}{4320--4333}.
\newblock
\urldef\tempurl%
\url{https://doi.org/10.18653/v1/2020.acl-main.398}
\showDOI{\tempurl}


\bibitem[\protect\citeauthoryear{Hu, Xu, Ye, Yan, Zhang, Zhang, Li, Zhang, Jin, Huang, and Zhou}{Hu et~al\mbox{.}}{2024}]%
        {DocOwl1.5}
\bibfield{author}{\bibinfo{person}{Anwen Hu}, \bibinfo{person}{Haiyang Xu}, \bibinfo{person}{Jiabo Ye}, \bibinfo{person}{Ming Yan}, \bibinfo{person}{Liang Zhang}, \bibinfo{person}{Bo Zhang}, \bibinfo{person}{Chen Li}, \bibinfo{person}{Ji Zhang}, \bibinfo{person}{Qin Jin}, \bibinfo{person}{Fei Huang}, {and} \bibinfo{person}{Jingren Zhou}.} \bibinfo{year}{2024}\natexlab{}.
\newblock \bibinfo{title}{mPLUG-DocOwl 1.5: Unified Structure Learning for OCR-free Document Understanding}.
\newblock
\newblock
\showeprint[arxiv]{2403.12895}~[cs.CV]
\urldef\tempurl%
\url{https://arxiv.org/abs/2403.12895}
\showURL{%
\tempurl}


\bibitem[\protect\citeauthoryear{Jauhar, Turney, and Hovy}{Jauhar et~al\mbox{.}}{2016}]%
        {SemiKnowledge}
\bibfield{author}{\bibinfo{person}{Sujay~Kumar Jauhar}, \bibinfo{person}{Peter Turney}, {and} \bibinfo{person}{Eduard Hovy}.} \bibinfo{year}{2016}\natexlab{}.
\newblock \showarticletitle{Tables as Semi-structured Knowledge for Question Answering}. In \bibinfo{booktitle}{\emph{Proceedings of the 54th Annual Meeting of the Association for Computational Linguistics (Volume 1: Long Papers)}}, \bibfield{editor}{\bibinfo{person}{Katrin Erk} {and} \bibinfo{person}{Noah~A. Smith}} (Eds.). \bibinfo{publisher}{Association for Computational Linguistics}, \bibinfo{address}{Berlin, Germany}, \bibinfo{pages}{474--483}.
\newblock
\urldef\tempurl%
\url{https://doi.org/10.18653/v1/P16-1045}
\showDOI{\tempurl}


\bibitem[\protect\citeauthoryear{Khot, Trivedi, Finlayson, Fu, Richardson, Clark, and Sabharwal}{Khot et~al\mbox{.}}{2023}]%
        {DecomposedPrompting}
\bibfield{author}{\bibinfo{person}{Tushar Khot}, \bibinfo{person}{Harsh Trivedi}, \bibinfo{person}{Matthew Finlayson}, \bibinfo{person}{Yao Fu}, \bibinfo{person}{Kyle Richardson}, \bibinfo{person}{Peter Clark}, {and} \bibinfo{person}{Ashish Sabharwal}.} \bibinfo{year}{2023}\natexlab{}.
\newblock \bibinfo{title}{Decomposed Prompting: A Modular Approach for Solving Complex Tasks}.
\newblock
\newblock
\showeprint[arxiv]{2210.02406}~[cs.CL]
\urldef\tempurl%
\url{https://arxiv.org/abs/2210.02406}
\showURL{%
\tempurl}


\bibitem[\protect\citeauthoryear{Li, He, Yashar, Cui, Ge, Zhang, Fainman, Zhang, and Chaudhuri}{Li et~al\mbox{.}}{2023}]%
        {Table-GPT}
\bibfield{author}{\bibinfo{person}{Peng Li}, \bibinfo{person}{Yeye He}, \bibinfo{person}{Dror Yashar}, \bibinfo{person}{Weiwei Cui}, \bibinfo{person}{Song Ge}, \bibinfo{person}{Haidong Zhang}, \bibinfo{person}{Danielle~Rifinski Fainman}, \bibinfo{person}{Dongmei Zhang}, {and} \bibinfo{person}{Surajit Chaudhuri}.} \bibinfo{year}{2023}\natexlab{}.
\newblock \bibinfo{title}{Table-GPT: Table-tuned GPT for Diverse Table Tasks}.
\newblock
\newblock
\showeprint[arxiv]{2310.09263}~[cs.CL]
\urldef\tempurl%
\url{https://arxiv.org/abs/2310.09263}
\showURL{%
\tempurl}


\bibitem[\protect\citeauthoryear{Li, Wang, Zhao, Yang, Du, Hu, Zhang, Ye, Li, Zhao, and Mao}{Li et~al\mbox{.}}{2024}]%
        {PETSQL}
\bibfield{author}{\bibinfo{person}{Zhishuai Li}, \bibinfo{person}{Xiang Wang}, \bibinfo{person}{Jingjing Zhao}, \bibinfo{person}{Sun Yang}, \bibinfo{person}{Guoqing Du}, \bibinfo{person}{Xiaoru Hu}, \bibinfo{person}{Bin Zhang}, \bibinfo{person}{Yuxiao Ye}, \bibinfo{person}{Ziyue Li}, \bibinfo{person}{Rui Zhao}, {and} \bibinfo{person}{Hangyu Mao}.} \bibinfo{year}{2024}\natexlab{}.
\newblock \bibinfo{title}{PET-SQL: A Prompt-Enhanced Two-Round Refinement of Text-to-SQL with Cross-consistency}.
\newblock
\newblock
\showeprint[arxiv]{2403.09732}~[cs.CL]
\urldef\tempurl%
\url{https://arxiv.org/abs/2403.09732}
\showURL{%
\tempurl}


\bibitem[\protect\citeauthoryear{Lin, Hasan, Kosalge, Cheung, and Parameswaran}{Lin et~al\mbox{.}}{2025}]%
        {TWIX}
\bibfield{author}{\bibinfo{person}{Yiming Lin}, \bibinfo{person}{Mawil Hasan}, \bibinfo{person}{Rohan Kosalge}, \bibinfo{person}{Alvin Cheung}, {and} \bibinfo{person}{Aditya~G. Parameswaran}.} \bibinfo{year}{2025}\natexlab{}.
\newblock \bibinfo{title}{TWIX: Automatically Reconstructing Structured Data from Templatized Documents}.
\newblock
\newblock
\showeprint[arxiv]{2501.06659}~[cs.DB]
\urldef\tempurl%
\url{https://arxiv.org/abs/2501.06659}
\showURL{%
\tempurl}


\bibitem[\protect\citeauthoryear{Liu, Li, Gong, Liu, Wu, Jiang, Liu, and Sun}{Liu et~al\mbox{.}}{2024}]%
        {GrabTable}
\bibfield{author}{\bibinfo{person}{Hao Liu}, \bibinfo{person}{Xin Li}, \bibinfo{person}{Mingming Gong}, \bibinfo{person}{Bing Liu}, \bibinfo{person}{Yunfei Wu}, \bibinfo{person}{Deqiang Jiang}, \bibinfo{person}{Yinsong Liu}, {and} \bibinfo{person}{Xing Sun}.} \bibinfo{year}{2024}\natexlab{}.
\newblock \showarticletitle{Grab What You Need: Rethinking Complex Table Structure Recognition with Flexible Components Deliberation}.
\newblock \bibinfo{journal}{\emph{Proceedings of the AAAI Conference on Artificial Intelligence}} \bibinfo{volume}{38}, \bibinfo{number}{4} (\bibinfo{date}{Mar.} \bibinfo{year}{2024}), \bibinfo{pages}{3603--3611}.
\newblock
\urldef\tempurl%
\url{https://doi.org/10.1609/aaai.v38i4.28149}
\showDOI{\tempurl}


\bibitem[\protect\citeauthoryear{Liu, Lin, Hewitt, Paranjape, Bevilacqua, Petroni, and Liang}{Liu et~al\mbox{.}}{2023}]%
        {lostinthemiddle}
\bibfield{author}{\bibinfo{person}{Nelson~F. Liu}, \bibinfo{person}{Kevin Lin}, \bibinfo{person}{John Hewitt}, \bibinfo{person}{Ashwin Paranjape}, \bibinfo{person}{Michele Bevilacqua}, \bibinfo{person}{Fabio Petroni}, {and} \bibinfo{person}{Percy Liang}.} \bibinfo{year}{2023}\natexlab{}.
\newblock \bibinfo{title}{Lost in the Middle: How Language Models Use Long Contexts}.
\newblock
\newblock
\showeprint[arxiv]{2307.03172}~[cs.CL]
\urldef\tempurl%
\url{https://arxiv.org/abs/2307.03172}
\showURL{%
\tempurl}


\bibitem[\protect\citeauthoryear{Lu, Song, Qin, Zhang, Zhang, and Wong}{Lu et~al\mbox{.}}{2025}]%
        {lu2025bridge}
\bibfield{author}{\bibinfo{person}{Jinwei Lu}, \bibinfo{person}{Yuanfeng Song}, \bibinfo{person}{Zhiqian Qin}, \bibinfo{person}{Haodi Zhang}, \bibinfo{person}{Chen Zhang}, {and} \bibinfo{person}{Raymond Chi-Wing Wong}.} \bibinfo{year}{2025}\natexlab{}.
\newblock \bibinfo{title}{Bridging the Gap: Enabling Natural Language Queries for NoSQL Databases through Text-to-NoSQL Translation}.
\newblock
\newblock
\showeprint[arxiv]{2502.11201}~[cs.DB]
\urldef\tempurl%
\url{https://arxiv.org/abs/2502.11201}
\showURL{%
\tempurl}


\bibitem[\protect\citeauthoryear{OpenAI}{OpenAI}{2024}]%
        {openai2024gpt4o}
\bibfield{author}{\bibinfo{person}{OpenAI}.} \bibinfo{year}{2024}\natexlab{}.
\newblock \bibinfo{title}{GPT-4o System Card}.
\newblock
\newblock
\showeprint[arxiv]{2410.21276}~[cs.CL]
\urldef\tempurl%
\url{https://arxiv.org/abs/2410.21276}
\showURL{%
\tempurl}


\bibitem[\protect\citeauthoryear{OpenAI, Achiam, Adler, Agarwal, Ahmad, Akkaya, et~al\mbox{.}}{OpenAI et~al\mbox{.}}{2024}]%
        {GPT4}
\bibfield{author}{\bibinfo{person}{OpenAI}, \bibinfo{person}{Josh Achiam}, \bibinfo{person}{Steven Adler}, \bibinfo{person}{Sandhini Agarwal}, \bibinfo{person}{Lama Ahmad}, \bibinfo{person}{Ilge Akkaya}, {et~al\mbox{.}}} \bibinfo{year}{2024}\natexlab{}.
\newblock \bibinfo{title}{GPT-4 Technical Report}.
\newblock
\newblock
\showeprint[arxiv]{2303.08774}~[cs.CL]
\urldef\tempurl%
\url{https://arxiv.org/abs/2303.08774}
\showURL{%
\tempurl}


\bibitem[\protect\citeauthoryear{Pasupat and Liang}{Pasupat and Liang}{2015a}]%
        {wikitq}
\bibfield{author}{\bibinfo{person}{Panupong Pasupat} {and} \bibinfo{person}{Percy Liang}.} \bibinfo{year}{2015}\natexlab{a}.
\newblock \showarticletitle{Compositional Semantic Parsing on Semi-Structured Tables}. In \bibinfo{booktitle}{\emph{Proceedings of the 53rd Annual Meeting of the Association for Computational Linguistics and the 7th International Joint Conference on Natural Language Processing (Volume 1: Long Papers)}}, \bibfield{editor}{\bibinfo{person}{Chengqing Zong} {and} \bibinfo{person}{Michael Strube}} (Eds.). \bibinfo{publisher}{Association for Computational Linguistics}, \bibinfo{address}{Beijing, China}, \bibinfo{pages}{1470--1480}.
\newblock
\urldef\tempurl%
\url{https://doi.org/10.3115/v1/P15-1142}
\showDOI{\tempurl}


\bibitem[\protect\citeauthoryear{Pasupat and Liang}{Pasupat and Liang}{2015b}]%
        {Compositional}
\bibfield{author}{\bibinfo{person}{Panupong Pasupat} {and} \bibinfo{person}{Percy Liang}.} \bibinfo{year}{2015}\natexlab{b}.
\newblock \showarticletitle{Compositional Semantic Parsing on Semi-Structured Tables}. In \bibinfo{booktitle}{\emph{Proceedings of the 53rd Annual Meeting of the Association for Computational Linguistics and the 7th International Joint Conference on Natural Language Processing (Volume 1: Long Papers)}}, \bibfield{editor}{\bibinfo{person}{Chengqing Zong} {and} \bibinfo{person}{Michael Strube}} (Eds.). \bibinfo{publisher}{Association for Computational Linguistics}, \bibinfo{address}{Beijing, China}, \bibinfo{pages}{1470--1480}.
\newblock
\urldef\tempurl%
\url{https://doi.org/10.3115/v1/P15-1142}
\showDOI{\tempurl}


\bibitem[\protect\citeauthoryear{Qin, Hui, Wang, Yang, Li, Li, Geng, Cao, Sun, Si, Huang, and Li}{Qin et~al\mbox{.}}{2022}]%
        {qin2022survey}
\bibfield{author}{\bibinfo{person}{Bowen Qin}, \bibinfo{person}{Binyuan Hui}, \bibinfo{person}{Lihan Wang}, \bibinfo{person}{Min Yang}, \bibinfo{person}{Jinyang Li}, \bibinfo{person}{Binhua Li}, \bibinfo{person}{Ruiying Geng}, \bibinfo{person}{Rongyu Cao}, \bibinfo{person}{Jian Sun}, \bibinfo{person}{Luo Si}, \bibinfo{person}{Fei Huang}, {and} \bibinfo{person}{Yongbin Li}.} \bibinfo{year}{2022}\natexlab{}.
\newblock \showarticletitle{A Survey on Text-to-SQL Parsing: Concepts, Methods, and Future Directions}.
\newblock \bibinfo{journal}{\emph{arXiv preprint arXiv:2208.13629}} (\bibinfo{year}{2022}).
\newblock
\urldef\tempurl%
\url{https://arxiv.org/abs/2208.13629}
\showURL{%
\tempurl}


\bibitem[\protect\citeauthoryear{Qu, Li, Li, Qin, Huo, Ma, and Cheng}{Qu et~al\mbox{.}}{2024}]%
        {BeforeGeneration}
\bibfield{author}{\bibinfo{person}{Ge Qu}, \bibinfo{person}{Jinyang Li}, \bibinfo{person}{Bowen Li}, \bibinfo{person}{Bowen Qin}, \bibinfo{person}{Nan Huo}, \bibinfo{person}{Chenhao Ma}, {and} \bibinfo{person}{Reynold Cheng}.} \bibinfo{year}{2024}\natexlab{}.
\newblock \showarticletitle{Before Generation, Align it! A Novel and Effective Strategy for Mitigating Hallucinations in Text-to-SQL Generation}. In \bibinfo{booktitle}{\emph{Findings of the Association for Computational Linguistics: ACL 2024}}. \bibinfo{publisher}{Association for Computational Linguistics}, \bibinfo{address}{Bangkok, Thailand}, \bibinfo{pages}{5456--5471}.
\newblock
\urldef\tempurl%
\url{https://doi.org/10.18653/v1/2024.findings-acl.324}
\showDOI{\tempurl}


\bibitem[\protect\citeauthoryear{Scholak, Schucher, and Bahdanau}{Scholak et~al\mbox{.}}{2021}]%
        {PICARD}
\bibfield{author}{\bibinfo{person}{Torsten Scholak}, \bibinfo{person}{Nathan Schucher}, {and} \bibinfo{person}{Dzmitry Bahdanau}.} \bibinfo{year}{2021}\natexlab{}.
\newblock \showarticletitle{{PICARD}: Parsing Incrementally for Constrained Auto-Regressive Decoding from Language Models}. In \bibinfo{booktitle}{\emph{Proceedings of the 2021 Conference on Empirical Methods in Natural Language Processing}}. \bibinfo{publisher}{Association for Computational Linguistics}, \bibinfo{address}{Online and Punta Cana, Dominican Republic}, \bibinfo{pages}{9895--9901}.
\newblock
\urldef\tempurl%
\url{https://doi.org/10.18653/v1/2021.emnlp-main.779}
\showDOI{\tempurl}


\bibitem[\protect\citeauthoryear{Schuster, Lelkes, Sun, Gupta, Berant, Cohen, and Metzler}{Schuster et~al\mbox{.}}{2024}]%
        {schuster2023semqa}
\bibfield{author}{\bibinfo{person}{Tal Schuster}, \bibinfo{person}{Adam Lelkes}, \bibinfo{person}{Haitian Sun}, \bibinfo{person}{Jai Gupta}, \bibinfo{person}{Jonathan Berant}, \bibinfo{person}{William Cohen}, {and} \bibinfo{person}{Donald Metzler}.} \bibinfo{year}{2024}\natexlab{}.
\newblock \showarticletitle{{SEMQA}: Semi-Extractive Multi-Source Question Answering}. In \bibinfo{booktitle}{\emph{Proceedings of the 2024 Conference of the North American Chapter of the Association for Computational Linguistics: Human Language Technologies (Volume 1: Long Papers)}}. \bibinfo{publisher}{Association for Computational Linguistics}, \bibinfo{address}{Mexico City, Mexico}, \bibinfo{pages}{1363--1381}.
\newblock
\urldef\tempurl%
\url{https://doi.org/10.18653/v1/2024.naacl-long.74}
\showDOI{\tempurl}


\bibitem[\protect\citeauthoryear{Sui, Zhou, Zhou, Han, and Zhang}{Sui et~al\mbox{.}}{2024}]%
        {sui2024table}
\bibfield{author}{\bibinfo{person}{Yuan Sui}, \bibinfo{person}{Mengyu Zhou}, \bibinfo{person}{Mingjie Zhou}, \bibinfo{person}{Shi Han}, {and} \bibinfo{person}{Dongmei Zhang}.} \bibinfo{year}{2024}\natexlab{}.
\newblock \showarticletitle{Table Meets LLM: Can Large Language Models Understand Structured Table Data? A Benchmark and Empirical Study}. In \bibinfo{booktitle}{\emph{Proceedings of the 17th ACM International Conference on Web Search and Data Mining (WSDM 2024)}}. \bibinfo{publisher}{Association for Computing Machinery}, \bibinfo{address}{Mérida, Mexico}, \bibinfo{pages}{123--134}.
\newblock
\urldef\tempurl%
\url{https://doi.org/10.1145/3616855.3635752}
\showDOI{\tempurl}


\bibitem[\protect\citeauthoryear{Wang, Zhang, Ma, Sun, Wang, and Wang}{Wang et~al\mbox{.}}{2018}]%
        {NeuralQuestion}
\bibfield{author}{\bibinfo{person}{Hao Wang}, \bibinfo{person}{Xiaodong Zhang}, \bibinfo{person}{Shuming Ma}, \bibinfo{person}{Xu Sun}, \bibinfo{person}{Houfeng Wang}, {and} \bibinfo{person}{Mengxiang Wang}.} \bibinfo{year}{2018}\natexlab{}.
\newblock \showarticletitle{A Neural Question Answering Model Based on Semi-Structured Tables}. In \bibinfo{booktitle}{\emph{Proceedings of the 27th International Conference on Computational Linguistics}}, \bibfield{editor}{\bibinfo{person}{Emily~M. Bender}, \bibinfo{person}{Leon Derczynski}, {and} \bibinfo{person}{Pierre Isabelle}} (Eds.). \bibinfo{publisher}{Association for Computational Linguistics}, \bibinfo{address}{Santa Fe, New Mexico, USA}, \bibinfo{pages}{1941--1951}.
\newblock
\urldef\tempurl%
\url{https://aclanthology.org/C18-1165/}
\showURL{%
\tempurl}


\bibitem[\protect\citeauthoryear{Wang, Yang, Huang, Yang, Majumder, and Wei}{Wang et~al\mbox{.}}{2024a}]%
        {MultilingualE5}
\bibfield{author}{\bibinfo{person}{Liang Wang}, \bibinfo{person}{Nan Yang}, \bibinfo{person}{Xiaolong Huang}, \bibinfo{person}{Linjun Yang}, \bibinfo{person}{Rangan Majumder}, {and} \bibinfo{person}{Furu Wei}.} \bibinfo{year}{2024}\natexlab{a}.
\newblock \bibinfo{title}{Multilingual E5 Text Embeddings: A Technical Report}.
\newblock
\newblock
\showeprint[arxiv]{2402.05672}~[cs.CL]
\urldef\tempurl%
\url{https://arxiv.org/abs/2402.05672}
\showURL{%
\tempurl}


\bibitem[\protect\citeauthoryear{Wang, Zhang, Nie, and Kim}{Wang et~al\mbox{.}}{2024b}]%
        {ToolAssisted}
\bibfield{author}{\bibinfo{person}{Zhongyuan Wang}, \bibinfo{person}{Richong Zhang}, \bibinfo{person}{Zhijie Nie}, {and} \bibinfo{person}{Jaein Kim}.} \bibinfo{year}{2024}\natexlab{b}.
\newblock \bibinfo{title}{Tool-Assisted Agent on SQL Inspection and Refinement in Real-World Scenarios}.
\newblock
\newblock
\showeprint[arxiv]{2408.16991}~[cs.CL]
\urldef\tempurl%
\url{https://arxiv.org/abs/2408.16991}
\showURL{%
\tempurl}


\bibitem[\protect\citeauthoryear{Xie, Xu, Zhao, and Guo}{Xie et~al\mbox{.}}{2025}]%
        {OpenSearchSQL}
\bibfield{author}{\bibinfo{person}{Xiangjin Xie}, \bibinfo{person}{Guangwei Xu}, \bibinfo{person}{Lingyan Zhao}, {and} \bibinfo{person}{Ruijie Guo}.} \bibinfo{year}{2025}\natexlab{}.
\newblock \bibinfo{title}{OpenSearch-SQL: Enhancing Text-to-SQL with Dynamic Few-shot and Consistency Alignment}.
\newblock
\newblock
\showeprint[arxiv]{2502.14913}~[cs.CL]
\urldef\tempurl%
\url{https://arxiv.org/abs/2502.14913}
\showURL{%
\tempurl}


\bibitem[\protect\citeauthoryear{Ye, Hui, Yang, Li, Huang, and Li}{Ye et~al\mbox{.}}{2023}]%
        {LLMDecomposers}
\bibfield{author}{\bibinfo{person}{Yunhu Ye}, \bibinfo{person}{Binyuan Hui}, \bibinfo{person}{Min Yang}, \bibinfo{person}{Binhua Li}, \bibinfo{person}{Fei Huang}, {and} \bibinfo{person}{Yongbin Li}.} \bibinfo{year}{2023}\natexlab{}.
\newblock \bibinfo{title}{Large Language Models are Versatile Decomposers: Decompose Evidence and Questions for Table-based Reasoning}.
\newblock
\newblock
\showeprint[arxiv]{2301.13808}~[cs.CL]
\urldef\tempurl%
\url{https://arxiv.org/abs/2301.13808}
\showURL{%
\tempurl}


\bibitem[\protect\citeauthoryear{Zan, Chen, Zhang, Lu, Wu, Guan, Wang, and Lou}{Zan et~al\mbox{.}}{2023}]%
        {zan2023nl2code}
\bibfield{author}{\bibinfo{person}{Daoguang Zan}, \bibinfo{person}{Bei Chen}, \bibinfo{person}{Fengji Zhang}, \bibinfo{person}{Dianjie Lu}, \bibinfo{person}{Bingchao Wu}, \bibinfo{person}{Bei Guan}, \bibinfo{person}{Yongji Wang}, {and} \bibinfo{person}{Jian-Guang Lou}.} \bibinfo{year}{2023}\natexlab{}.
\newblock \showarticletitle{Large Language Models Meet NL2Code: A Survey}. In \bibinfo{booktitle}{\emph{Proceedings of the 61st Annual Meeting of the Association for Computational Linguistics (Volume 1: Long Papers)}}. \bibinfo{publisher}{Association for Computational Linguistics}, \bibinfo{address}{Toronto, Canada}, \bibinfo{pages}{7443--7464}.
\newblock
\urldef\tempurl%
\url{https://doi.org/10.18653/v1/2023.acl-long.411}
\showDOI{\tempurl}


\bibitem[\protect\citeauthoryear{Zhang, Yue, Li, and Sun}{Zhang et~al\mbox{.}}{2024}]%
        {TableLLaMA}
\bibfield{author}{\bibinfo{person}{Tianshu Zhang}, \bibinfo{person}{Xiang Yue}, \bibinfo{person}{Yifei Li}, {and} \bibinfo{person}{Huan Sun}.} \bibinfo{year}{2024}\natexlab{}.
\newblock \showarticletitle{TableLlama: Towards Open Large Generalist Models for Tables}. In \bibinfo{booktitle}{\emph{Proceedings of the 2024 Conference of the North American Chapter of the Association for Computational Linguistics: Human Language Technologies (Volume 1: Long Papers)}}. \bibinfo{publisher}{Association for Computational Linguistics}, \bibinfo{address}{Mexico City, Mexico}, \bibinfo{pages}{6024--6044}.
\newblock
\urldef\tempurl%
\url{https://doi.org/10.18653/v1/2024.naacl-long.335}
\showDOI{\tempurl}


\bibitem[\protect\citeauthoryear{Zhang, Luo, Zhang, Ma, Zhang, Li, Li, Yao, Xu, Zhou, Zhang-Li, Yu, Zhao, Li, and Tang}{Zhang et~al\mbox{.}}{2025}]%
        {TableLLM}
\bibfield{author}{\bibinfo{person}{Xiaokang Zhang}, \bibinfo{person}{Sijia Luo}, \bibinfo{person}{Bohan Zhang}, \bibinfo{person}{Zeyao Ma}, \bibinfo{person}{Jing Zhang}, \bibinfo{person}{Yang Li}, \bibinfo{person}{Guanlin Li}, \bibinfo{person}{Zijun Yao}, \bibinfo{person}{Kangli Xu}, \bibinfo{person}{Jinchang Zhou}, \bibinfo{person}{Daniel Zhang-Li}, \bibinfo{person}{Jifan Yu}, \bibinfo{person}{Shu Zhao}, \bibinfo{person}{Juanzi Li}, {and} \bibinfo{person}{Jie Tang}.} \bibinfo{year}{2025}\natexlab{}.
\newblock \bibinfo{title}{TableLLM: Enabling Tabular Data Manipulation by LLMs in Real Office Usage Scenarios}.
\newblock
\newblock
\showeprint[arxiv]{2403.19318}~[cs.CL]
\urldef\tempurl%
\url{https://arxiv.org/abs/2403.19318}
\showURL{%
\tempurl}


\bibitem[\protect\citeauthoryear{Zhang, Henkel, Floratou, Cahoon, Deep, and Patel}{Zhang et~al\mbox{.}}{2023}]%
        {ReAcTable}
\bibfield{author}{\bibinfo{person}{Yunjia Zhang}, \bibinfo{person}{Jordan Henkel}, \bibinfo{person}{Avrilia Floratou}, \bibinfo{person}{Joyce Cahoon}, \bibinfo{person}{Shaleen Deep}, {and} \bibinfo{person}{Jignesh~M. Patel}.} \bibinfo{year}{2023}\natexlab{}.
\newblock \bibinfo{title}{ReAcTable: Enhancing ReAct for Table Question Answering}.
\newblock
\newblock
\showeprint[arxiv]{2310.00815}~[cs.DB]
\urldef\tempurl%
\url{https://arxiv.org/abs/2310.00815}
\showURL{%
\tempurl}


\bibitem[\protect\citeauthoryear{Zheng, Feng, Si, She, Lin, Jiang, and Wang}{Zheng et~al\mbox{.}}{2024}]%
        {TableLLaVA}
\bibfield{author}{\bibinfo{person}{Mingyu Zheng}, \bibinfo{person}{Xinwei Feng}, \bibinfo{person}{Qingyi Si}, \bibinfo{person}{Qiaoqiao She}, \bibinfo{person}{Zheng Lin}, \bibinfo{person}{Wenbin Jiang}, {and} \bibinfo{person}{Weiping Wang}.} \bibinfo{year}{2024}\natexlab{}.
\newblock \bibinfo{title}{Multimodal Table Understanding}.
\newblock
\newblock
\showeprint[arxiv]{2406.08100}~[cs.CL]
\urldef\tempurl%
\url{https://arxiv.org/abs/2406.08100}
\showURL{%
\tempurl}


\bibitem[\protect\citeauthoryear{Zhu, Liu, Feng, Wang, Li, and Chua}{Zhu et~al\mbox{.}}{2024}]%
        {TATLLM}
\bibfield{author}{\bibinfo{person}{Fengbin Zhu}, \bibinfo{person}{Ziyang Liu}, \bibinfo{person}{Fuli Feng}, \bibinfo{person}{Chao Wang}, \bibinfo{person}{Moxin Li}, {and} \bibinfo{person}{Tat-Seng Chua}.} \bibinfo{year}{2024}\natexlab{}.
\newblock \bibinfo{title}{TAT-LLM: A Specialized Language Model for Discrete Reasoning over Tabular and Textual Data}.
\newblock
\newblock
\showeprint[arxiv]{2401.13223}~[cs.CL]
\urldef\tempurl%
\url{https://arxiv.org/abs/2401.13223}
\showURL{%
\tempurl}


\end{thebibliography}

\end{document}